\documentclass{article} 
\usepackage{main,times}

\usepackage{notations}

\usepackage{xcolor}
\definecolor{darkblue}{rgb}{0,0,0.5}
\definecolor{firebrick}{rgb}{0.75,0.125,0.125}
\definecolor{darkgreen}{rgb}{0,0.5,0}
\definecolor{light-gray}{gray}{0.5}

\usepackage[colorlinks=true,linkcolor=firebrick,citecolor=darkgreen,urlcolor=darkblue]{hyperref}
\usepackage{url}
\usepackage{units}

\usepackage[]{algorithm2e}
\usepackage{wrapfig}
\usepackage{graphicx}
\usepackage{subfig}
\usepackage{comment}
\usepackage{amsmath}
\usepackage{amssymb}

\usepackage{booktabs}
\usepackage{multirow}

\title{Multi-timestep models for Model-based Reinforcement Learning}

\author{%
Abdelhakim Benechehab\thanks{Noah's Ark Lab, Huawei Technologies France. Correspondence to: Abdelhakim Benechehab $<$abdelhakim.benechehab1@huawei.com$>$.}\\
\And
Giuseppe Paolo\footnotemark[1]\\
\And
Albert Thomas\footnotemark[1]\\
\And
Maurizio Filippone\thanks{Department of Data Science, EURECOM, France.}\\
\And
Bal\'{a}zs K\'{e}gl\footnotemark[1]\\
}

\usepackage{xcolor}
\usepackage[commandnameprefix=always]{changes}

\definecolor{color-giuseppe}{rgb}{0.2, .8, 0.2}
\definechangesauthor[name={Giuseppe}, color=color-giuseppe]{G}

\definechangesauthor[name={Maurizio}, color=blue]{M}

\definechangesauthor[name={Albert}, color=orange]{A}

\definechangesauthor[name={Balazs}, color=cyan]{B}

\iclrfinalcopy 
\begin{document}

\maketitle

\begin{abstract}
In model-based reinforcement learning (MBRL), most algorithms rely on simulating trajectories from one-step dynamics models learned on data. A critical challenge of this approach is the compounding of one-step prediction errors as length of the trajectory grows. In this paper we tackle this issue by using a multi-timestep objective to train one-step models. Our objective is a weighted sum of a loss function (e.g., negative log-likelihood) at various future horizons. We explore and test a range of weights profiles. We find that exponentially decaying weights lead to models that significantly improve the long-horizon R2 score. This improvement is particularly noticeable when the models were evaluated on noisy data. Finally, using a soft actor-critic (SAC) agent in pure batch reinforcement learning (RL) and iterated batch RL scenarios, we found that our multi-timestep models outperform or match standard one-step models. This was especially evident in a noisy variant of the considered environment, highlighting the potential of our approach in real-world applications.
\end{abstract}

\section{Introduction}
\label{secIntroduction}

Reinforcement learning is a paradigm where a control agent (or policy) learns through interacting with a dynamic system (or environment) and receives feedback in the form of rewards. This approach has proven successful in addressing some of the most challenging problems globally, as evidenced by \citet{SilveR2017, SilveR2018, Mnih2015, Vinyals2019}. However, reinforcement learning remains largely confined to simulated environments and does not extend to real-world engineering systems. This limitation is primarily due to two factors: i) the scarcity of data resulting from operational constraints, and ii) the safety standards associated with these systems. Model-based reinforcement learning (MBRL) is an approach that can potentially narrow the gap between RL and applications.

Model-based reinforcement learning (MBRL) algorithms alternate two steps: i) model learning, a supervised learning problem to learn the dynamics of the environment, and ii) policy optimization, where a policy and/or a value function is learned by sampling from the learned dynamics. MBRL is recognized for its sample efficiency, as the learning of policy/value is conducted (either wholly or partially) from imaginary model rollouts (also referred to as background planning), which are more cost-effective and readily available than rollouts in the true dynamics \citep{JanneR2019}. Moreover, a predictive model that performs well out-of-distribution facilitates easy transfer of the true model to new tasks or areas not included in the model training dataset \citep{Yu2020}.

While model-based reinforcement learning (MBRL) algorithms have achieved significant success, they are prone to \emph{compounding errors} when planning over extended horizons \citep{Lambert_2022}. This issue arises due to the propagation of model bias, leading to highly inaccurate predictions at longer horizons. This can be problematic in real-world applications, as it can result in out-of-distribution states that may violate the physical or safety constraints of the environment. The root of compounding errors is in the nature of the models used in MBRL. Typically, these are one-step  models that predict the next state based on the current state and executed action. Long rollouts are then generated by iteratively applying these models, leading to compunding errors. To address this issue, we propose aligning the training objective of these models to optimize the long-horizon error.

Our key contributions are the following:

\begin{itemize}
    \item We propose a novel training objective for one-step predictive models. It consists in a weighted sum of the loss function at multiple future horizons.
    \item We perform an ablation study on different strategies to weigh the loss function at multiple future horizons, finding that exponentially decaying weights lead to the best results.
    \item Finally, we evaluate the multi-timestep models in pure batch and iterated batch MBRL.
\end{itemize}

\section{Related Work}

\paragraph{Model-based reinforcement learning} MBRL has been effectively used in iterated batch RL by alternating between model learning and planning \citep{Deisenroth2011,HafneR2021,Gal2016,Levine2013,Chua2018,JanneR2019,Kegl2021}, and in the offline (pure batch) RL where we do one step of model learning followed by policy learning \citep{Yu2020,Kidambi2020,Lee2021,Argenson2021,Zhan2021,Yu2021,Liu2021,Benechehab_2023}. 
Planning is used either at decision time via model-predictive control (MPC) \citep{Draeger1995,Chua2018,HafneR2019,Pinneri2020,Kegl2021}, or in the background where a model-free agent is learned on imagined model rollouts (Dyna; \citet{JanneR2019,Sutton1991,Sutton1992,Ha2018}), or both. 
For example, model-based policy optimization (MBPO) \citep{JanneR2019} trains an ensemble of feed-forward models and generates imaginary rollouts to train a soft actor-critic agent.

\paragraph{Multi-step predictions} Multi-timestep dynamics modeling was referred to in early works about temporal abstraction \citep{Sutton1999, Precup1998} and mixture of timescales models in tabular MDPs \citep{Precup1997, Singh1992, Sutton1985, sutton1995}. It is also partially addressed in the recent MBRL literature by means of recurrent state space models \citep{Hochreiter_1997, Chung_2014, Ha2018, HafneR2019, HafneR2021, Silver_2017}. However, these methods rely on heavy computation and approximated inference schemes such as variational inference. The problem of multi-step forecasting was also addressed in the time series literature \citep{Bentaieb_2012a, Bentaieb_2012b, Venkatraman_2015, Chandra_2021}, but none directly aligns the single-step forecasting model training objective to match the multi-step prediction objective, as we propose.

\paragraph{Fixed/variable-horizon architectures} Regarding fixed-horizon models, \cite{Asadi_2018, Asadi_2019} designed a model that has a distinct neural block for each prediction step, which reduces error accumulation but enlarges the model. \cite{Whitney_2019} demonstrated that a simple fixed-horizon model has better accuracy and can be used for planning with CEM. 
However, they had to change the reward function since they did not have access to all the visited states when planning with a fixed-horizon model (discussed in section \ref{secFixedHor}). 
For multi-horizon architectures, \cite{Lambert_2021} proposed a model that takes the prediction horizon and the parameters of the current policy as input, instead of the full action sequence. Meanwhile, \cite{Whitney_2019} used a dynamic-horizon architecture by padding the shorter action sequences with zeros to fit the input size constraint.

\paragraph{Compounding errors in MBRL} 
\cite{Lambert_2022} showed that the stability of the physical system affects the error profile of a model. 
\cite{Talvitie_2014} proposed \emph{hallucinated replay}, which inputs to the model its own generated (noisy) outputs. 
This technique teaches the model to recover from its own errors (similar to denoising), and produces models that have lower one-step accuracy but higher returns in some scenarios. 
A similar idea (multi-step error) is used by \cite{Byravan_2021} to train models in the context of MPC, and by \cite{Venkatraman_2015} who casts the error-recovery task as imitation learning. 
In the same vein,  \cite{Xiao_2023} learned a criterion based on the $h$-step error of the model, and adjusted the planning horizon $h$ accordingly.



\section{Preliminaries}

The conventional framework used in RL is the finite-horizon \textbf{Markov decision process (MDP)}, denoted as $\mathcal{M} = \langle  \mathcal{S}, \mathcal{A}, p, r, \rho_0, \gamma \rangle$. 
In this notation, $\mathcal{S}$ is the state space and $\mathcal{A}$ is the action space. 
The transition dynamics, which could be stochastic, are represented by $p : \mathcal{S} \times \mathcal{A} \leadsto \mathcal{S}$. The reward function is denoted by $r : \mathcal{S} \times \mathcal{A} \rightarrow \mathbb{R}$. The initial state distribution is given by $\rho_0$, and the discount factor is represented by $\gamma$, and lies within the range of [0,1].

The objective of RL is to identify a policy $\pi(s)$ which is a distribution over the action space ($\mathcal{A}$) for each state ($s \in \mathcal{S}$). This policy aims at maximizing the expected sum of discounted rewards, denoted as $J(\pi, \mathcal{M}) := \mathbb{E}_{s_0 \sim \rho_0, a_t \sim \pi, \, s_{t>0} \sim p}[ \sum_{t=0}^{H} \gamma^t r(s_t, a_t) ]$, where $H$ represents the horizon of the MDP.
Under a given policy $\pi$, the state-action value function (also known as the Q-function) at a specific $(s,a) \in \mathcal{S} \times \mathcal{A}$ pair is defined as the expected sum of discounted rewards. This expectation starts from state $s$, takes action $a$, and follows the policy $\pi$ until termination: $ Q^\pi(s, a) = \mathbb{E}_{a_{t>0} \sim \pi,  s_{t>0} \sim p} \big[ \sum_{t=0}^H \gamma^{t} r(s_t,a_t) \mid s_0=s, a_0=a \big]$. Similarly, we can define the state value function by taking the expectation with respect to the initial action $a_0$: $ V^\pi(s) = \mathbb{E}_{a_t \sim \pi,  s_{t>0} \sim p} \big[ \sum_{t=0}^H \gamma^{t} r(s_t,a_t) \mid s_0=s\big]$.

\textbf{Model-based RL (MBRL)} algorithms address the supervised learning challenge of estimating the dynamics of the environment $\hat{p}$ (and sometimes also the reward function $\hat{r}$) from data collected when interacting with the real system. The loss function is typically the log-likelihood $\mathcal{L}(\mathcal{D}; \hat{p}) = \frac{1}{N} \sum_{i=1}^N \log  \hat{p}(s^i_{t+1}|s^i_t, a^i_t)$ or Mean Squared Error (MSE) for deterministic models. 
The learned model can subsequently be employed for policy search under the MDP $\widehat{\mathcal{M}} = \langle  \mathcal{S}, \mathcal{A}, \hat{p}, r, \rho_0, \gamma \rangle$.
This MDP shares the state and action spaces $\mathcal{S}, \mathcal{A}$, reward function $r$, with the true environment $\mathcal{M}$, but learns the transition probability $\hat{p}$ from the dataset $\mathcal{D}$. The policy $\hat{\pi} = \argmax_\pi J(\pi, \widehat{\mathcal{M}})$ learned on $\widehat{\mathcal{M}}$ is not guaranteed to be optimal under the true MDP $\mathcal{M}$ due to distribution shift and model bias. 

In \textbf{pure batch (or offline) RL}, we are given a set of $N$ transitions $\mathcal{D}=\{(s_t^i,a_t^i,r_t^i,s_{t+1}^i)\}_{i=1}^N$. These transitions are generated by an unknown behavioral policy, $\pi^\beta$. 
The challenge in pure batch RL is to learn a good policy in a single shot, without further interacting with the environment $\mathcal{M}$, even though our objective is to optimize $J(\pi, \mathcal{M})$ with $\pi \not= \pi^\beta$.

Similarly to \citet{Chua2018, Kegl2021}, we learn a probabilistic dynamics model with a prediction and an uncertainty on the prediction. Formally, $s_{t+1} \rLeadsto \hat{p}_\theta(s_t,a_t) = \mathcal{N}\big(\mu_\theta( s_t,a_t),\sigma_\theta( s_t,a_t)\big)$, where $\mathcal{N}$ is a multivariate Gaussian, and $\theta$ is the learned parameters of the predictive model. In practical applications, fully connected neural networks are often employed due to their proven capabilities as powerful function approximators \citep{Nagabandi2018,Chua2018, Kegl2021}, and their suitability for high-dimensional environments over simpler nonparametric models such as Gaussian processes. Following previous research \citep{Chua2018, Kegl2021}, we assume a diagonal covariance matrix for which we learn the logarithm of the diagonal entries $\sigma_\theta = \mathrm{Diag}(\exp(l_\theta))$, where $l_\theta$ denotes the output of the neural network.

\section{Methods}

\subsection{Problem statement}

In MBRL it is common to use a model $\hat{p}$ that predicts the state one-step ahead $s_{t+1} \rLeadsto \hat{p}(s_t, a_t)$.
We train this model to optimize the one-step predictive error $L\big(s_{t+1}, \hat{p}(s_t, a_t)\big)$ (MSE or NLL for stochastic modeling) in a supervised learning setting. To learn a policy, we use these models for planning $h$ steps ahead by applying a procedure called \emph{rollout}: we generate $s_{t+j} \rLeadsto\hat{p}(s_{t+j-1}, a_{t+j-1})$ recursively for $j = 1, \ldots, h$. Here $(a_t, \ldots, a_{t+h-1}) = \ba_{t:t+h}$ is either a fixed action sequence generated by planning or sampling policy $a_{t+j} \rLeadsto \pi(s_{t+j})$ for $j = 1, \ldots, h$, on the fly. Formally, let
\begin{eqnarray}\label{eqnRecursive}
\hat{p}^1(s_t, a_t) & = & \hat{p}(s_t, a_t)  \text{ and } \\ \nonumber\hat{p}^j(s_t, \ba_{t:t+j}) & = &\hat{p}\big(\hat{p}^{j-1}(s_t, \ba_{t:t+j-1}), a_{t+j-1}\big) \text{ for } j = 2, \ldots, h.
\end{eqnarray}

Using $\hat{p}^h(s_t, \ba_{t:t+h})$ to estimate $s_{t+h}$
is problematic for two reasons:
\begin{itemize}
    \item A distribution mismatch occurs between the inputs that the model was trained on (sampled from the true unknown transition distribution) and the inputs the model being evaluated on (sampled from the predictive distribution of the model; \cite{Talvitie_2014, Talvitie_2017}).
    \item The predictive error (and the modeled uncertainty in the case of stochastic models) will propagate through the successive model calls, leading to compounding errors \citep{Lambert_2022, Talvitie_2014, Venkatraman_2015}.
\end{itemize}


To mitigate these issues, we propose  models that learn to predict the state $s_{t+h}$ after an arbitrary future horizon $h$ given $s_t$ and an action sequence $\ba_{t:t+h} $. We identify two levels in addressing the stated problem:
\begin{enumerate}
\item First we build \emph{direct} models $\hat{p}_h(s_t, \ba_{t:t+h})$ that predict the state at a fixed future horizon $h$ (similar to \cite{Whitney_2019}). These models cannot directly be used for planning since they do not give access to intermediate rewards $(r_{t+1},\ldots,r_{t+h-1})$, but they serve to motivate the second level, our \textbf{main contribution}:
\item We build one-step models $\hat{p}$ but train them to predict the $h$-step error (\ref{eqnHStep}) using the recursive formula (\ref{eqnRecursive}). Since $\hat{p}$ is a neural net, here we are essentially using recurrent nets.
\end{enumerate}

We start our analysis from Level~1 in Section~\ref{secFixedHor}, and address Level~2 in Section~\ref{secImprove1}. Finally, the experimental setup and the performance evaluation of the agents and our proposed models is showcased in Section~\ref{secExp}.

\subsection{Fixed-horizon models}
\label{secFixedHor}

Fixed-horizon models $\hat{p}_h$ take as input the current state $s_t$ and an action sequence $\ba_{t:t+h} = (a_t,\ldots,a_{t+h-1})$, and predict the parameters of a diagonal multi-variate Gaussian distribution over the state observed $h$ steps into the future: $\hat{s}_{t+h} \sim \mathcal{N}(\hat{\mu}_\theta(s_t,\ba_{t:t+h}), \hat{\sigma}_\theta(s_t,\ba_{t:t+h}))$ with $\theta$ representing the parameters of $\hat{p}_h$. 

\begin{figure}[htp]
\centering
\includegraphics[width=.475\textwidth]{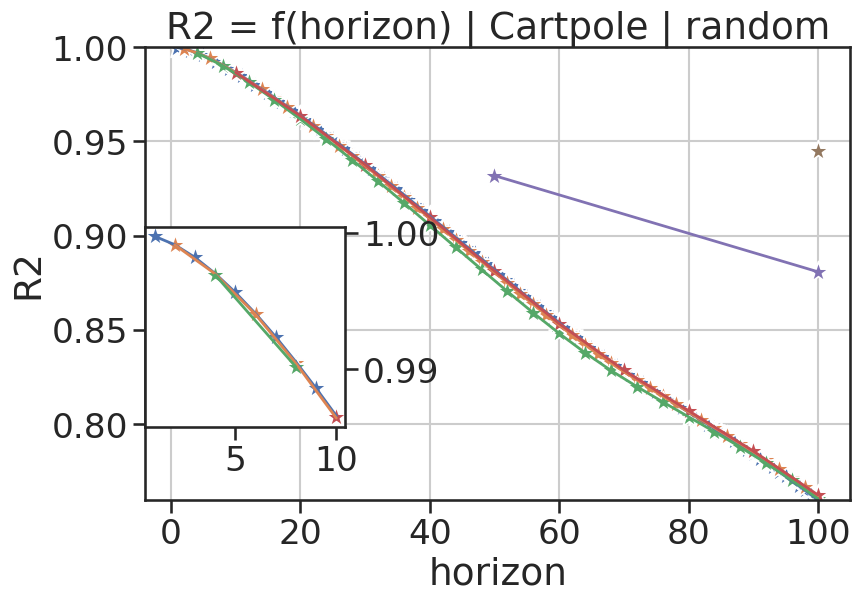}
\includegraphics[width=.475\textwidth]{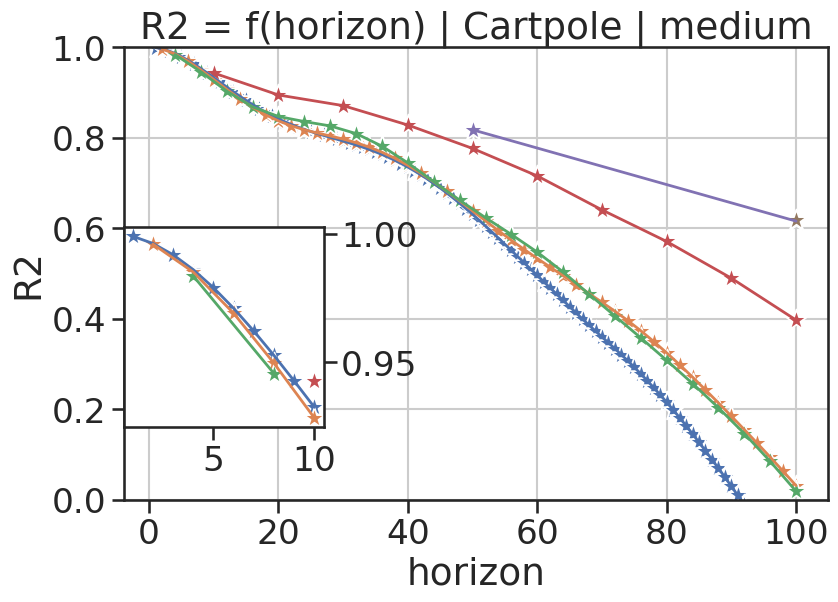}
\includegraphics[width=.8\textwidth]{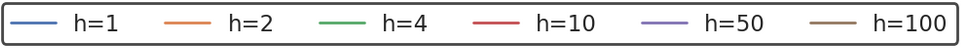}
\caption{R2 score (explained variance) as a function of the prediction horizon for several multi-step models. We use prediction horizons $h \in \{1,2,4,10,50,100\}$, and apply the predictors recursively according to (\ref{eqnRecursive}) up to maximum horizon $H = 100$. The difference between the two plots is the evaluation dataset: \emph{random} (where the behaviour policy is a random policy) and \emph{medium} (where the behaviour policy is a half-expert policy). A discussion on the differences between these datasets is given in Appendix~\ref{appendixdatasets}.}
\label{fig:fixed}
\end{figure}

Figure \ref{fig:fixed} shows the error at different horizons for different fixed-horizon models and datasets. For each horizon $h$, the error is computed by considering all sub-trajectories of size $h$ from the test set. We call each $h$-step model $\hat{p}_h$ recursively $H/h$ times to compute the prediction $\hat{p}_h^{H/h}$ up to $H$ steps ahead. 

In general, for a given horizon, we observe that the R2 score (explained variance) increases as we go from the one-step model to the hundred-step model. For the Cartpole random dataset, the short-horizon models are almost identical, while the fifty- and the hundred-step models achieve distinguishable R2 scores. In the medium dataset the improvement grows steadily with $h$.

\subsection{Training the one-step models with the error $h$ steps ahead}
\label{secImprove1}

\begin{figure}[htp]
\centering
\includegraphics[width=.475\textwidth]{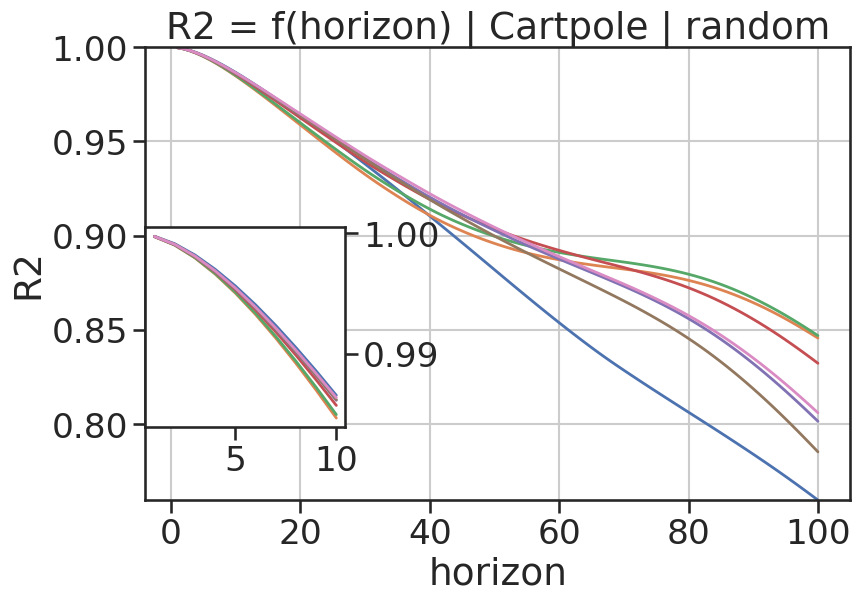}
\includegraphics[width=.475\textwidth]{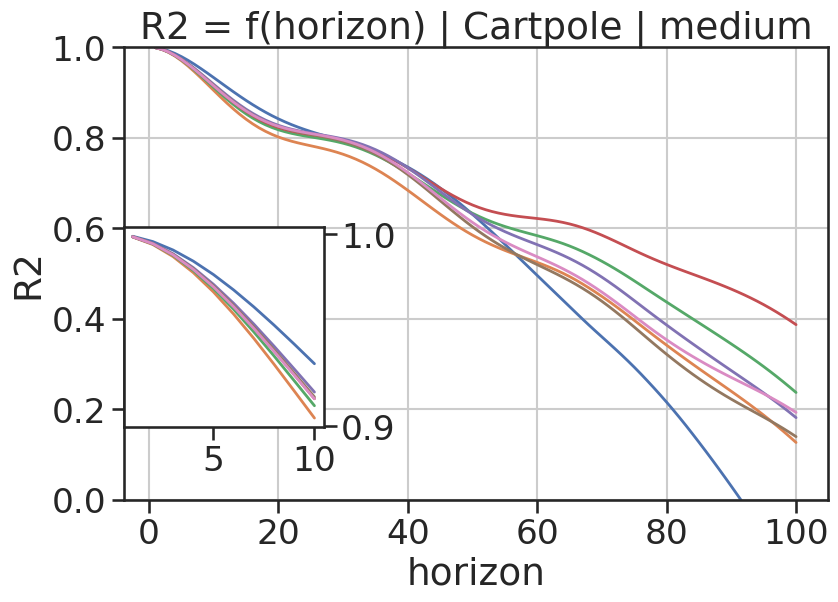}
\includegraphics[width=.8\textwidth]{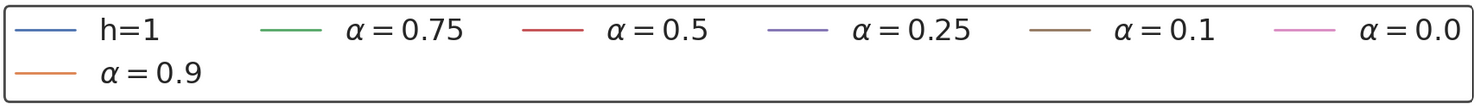}
\caption{R2 score (explained variance) as a function of the prediction horizon for different weight vectors of $\balpha = [\alpha, 1-\alpha]$. $\hat{p}$ is learned minimizing $\alpha \times L\big(s_{t+1},\hat{p}(s_t, a_t)\big) + (1 - \alpha) \times L\Big(s_{t+2},\hat{p}\big(\hat{p}(s_t, a_t), a_{t+1}\big) \Big)$.}
\label{fig:soft_R2}
\end{figure}

Figure~\ref{fig:fixed} shows that it may be worthy to train models for longer horizons. But, since our goal is to plan with an arbitrary horizon, we wanted to \emph{improve} the one-step predictor $\hat{p}$ rather than throwing it away. An obvious solution was to train the recurrent predictor $\hat{p}^h$ to minimize $L\big(s_{t+h},\hat{p}^h(s_t, \ba_{t:t+h})\big)$, but we found that this made the one-step prediction worse, even for the minimal case of $h=2$ (Figure~\ref{fig:soft_R2} with $\alpha=0$). As can be seen from the same figure, in both medium and random datasets the models where $\alpha < 1.0$, suffer a worse predictive error for short horizons, but recover and beat the vanilla one-step model down-the-horizon.


Following the empirical findings in the previous simple case, we hypothesize that the solution is to force the predictor $\hat{p}$ to be good on \emph{all} horizons. Formally, we define horizon-dependent weights $\balpha = (\alpha_1, \ldots, \alpha_h)$, and minimize the loss

\begin{equation} \label{eqnHStep}
    L_\balpha\big(\bs_{t+1:t+h+1},\hat{p}^{1:h}(s_t, \ba_{\cdot})\big) = \sum_{j=1}^h \alpha_j L\big(s_{t+j},\hat{p}^j(s_t, \ba_{t:t+j})\big).
\end{equation}

To proceed with \textit{gradient descent}-based optimization, we only need the gradient of the loss with respect to the model's parameters. Therefore, we provide an analysis of the analytical gradient of this loss in Appendix \ref{AppSecGrad}. We derive the gradient in equation \ref{eq5} to highlight the fact that unlike existing literature (hallucinated replay \cite{Talvitie_2014} and multi-step loss \cite{Byravan_2021}), our proposed method consists in back-propagating the gradient of the loss through the successive compositions of the model ($\hat{p}^{1:h}(s_t, \ba_{\cdot})$) figuring in $L_\balpha$. 
Furthermore, we find that the derivative of the generalized loss $L_\balpha$  can be expressed as a linear function of the derivative of the loss one-step ahead $L\big(s_{t+1}, \hat{p}(s_t, a_t)\big)$. This opens the door for gradient approximation ideas that are briefly discussed in Appendix \ref{AppSecGrad}, and intended to be explored in future follow-up works. 

An important matter that remains to solve is the choice of the weights $\{\alpha_j\}_{j \in \{1,\hdots,h\}}$. For instance, Figure~\ref{fig:soft_R2} shows that for $h=2$ the best weight vector $\balpha = (\alpha, 1-\alpha)$ depends on both the horizon and the dataset, and that it is neither $[1, 0]$ (the classical $\hat{p}$ trained for one-step error), nor $[0, 1]$ ($\hat{p}$ trained to minimize $L\Big(s_{t+2},\hat{p}\big(\hat{p}(s_t, a_t), a_{t+1}\big) \Big)$). In the next section, we present multiple choices for the weights $\alpha_j$, along with the intuition behind each one of them.

\subsubsection{How to choose $\{\alpha_j\}_{j \in \{1,\hdots,h\}}$ ?}

The choice of $\alpha_j$ is very important in this context as it allows us to balance the optimization problem between losses at different prediction horizons. In fact, the different loss terms are at different scales since the error grows with the prediction horizon, and therefore a sensible choice of these weights is critical.

\begin{figure}[ht]
\begin{center}
   \includegraphics[width=.55\linewidth]{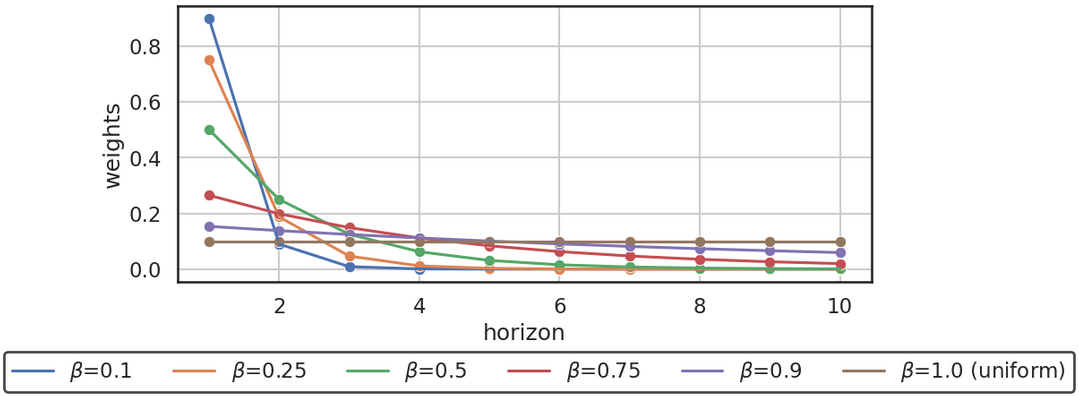}
\end{center}
\caption{For a maximum horizon of $10$, the figure shows the different weight profiles that we get by varying the decay parameter $\beta$.}
\label{fig:w}
\end{figure}

\begin{itemize}
    \item \textbf{Uniform.} $\alpha_j = 1/h$ The most intuitive way to handle the weighting of the losses is by using uniform weights. However, this method can be sub-optimal since the loss terms $L\big(s_{t+j},\hat{p}^j(s_t, \ba_{t:t+j})\big)$ grow with $j$ so in a flat sum long horizon errors will dominate.
    \item \textbf{Decay.} $\alpha_j = (\frac{1-\beta}{1-\beta^{h+1}}) \beta^j$ This method implements an exponential decay through a parameter $\beta$ (Figure \ref{fig:w} shows the weight profiles obtained using different values of $\beta$).
    \item \textbf{Learn.} $ \alpha_j = learnable $ In this setting, the $\alpha_j$ are left as free-parameters to be learned by the model. The expected outcome in this case is that the model puts the biggest weight on the one-step error $L\big(s_{t+1}, \hat{p}(s_t, a_t)\big)$ as it's smaller than further horizons. 
    \item \textbf{Proportional.} $\alpha_j \sim \frac{1}{L\big(s_{t+j},\hat{p}^j(s_t, \ba_{t:t+j})\big)}$ This method normalizes all the loss terms with their amplitude so that they become equivalent optimization-wise. 
\end{itemize}

\section{Experiments \& Results}
\label{secExp}

This section starts by presenting the experimental setup, followed by the experiments we conducted to evaluate the proposed models.

\subsection{Experimental setup}

We are interested in testing our models in two ways: ``static'' model learning from a fixed data set, and ``dynamic'' agent learning from model-generated traces.
While we are interested in building models that improve the predictive error beyond the one-step transition, the relationship between this surrogate metric and the return of the underlying agent is not trivial. Indeed, in many scenarios, models that have lower predictive ability lead to better performing agents \citep{Talvitie_2014}. 

Therefore, we suggest the following experimental settings:

\begin{itemize}
    \item \textbf{Pure batch (offline) RL.} In this setting, we dispose of multiple datasets with different characteristics. We then test the predictive error of our models against these datasets in a supervised-learning fashion. The main metric in this setting is the R2 score that we evaluate at different future horizons. The resulting models can then be used to train an RL agent (We use Soft-Actor Critic, implementation details can be found in Appendix~\ref{appendixsecImpDetails}) in what can be seen as a single-iteration MBRL.
    \item \textbf{Iterated batch RL.} In this setting, we alternate between generating a system trace (one episode from the real system), training the model using the traces collected so-far, and training the policy (RL agent) on the freshly updated model. We measure dynamic metrics such as the convergence pace and the mean asymptotic reward or return.
\end{itemize}

We conduct our experiments in the continuous control environment \textit{Cartpole swing-up}. We use the implementation of Deepmind Control \citep{Tassa_2018} which is based on the Mujoco physics simulator \citep{Todorov2012}. A description of this environment can be found in Appendix~\ref{appendixcartpole}. A detailed comparison of the static datasets used in the pure batch RL setting is given in Appendix~\ref{appendixdatasets}.

\paragraph{Noisy Environment} In an effort to further examine the applicability of multi-timestep models, we suggest the use of a \textit{Cartpole swing-up} variant that is characterized by the addition of Gaussian noise with a fixed standard deviation (equal to $1\%$ of the range of a given variable) to all observable variables. The Noisy Cartpole environment is of particular interest due to the potential for multi-timestep models to experience significant failures as a result of the accumulation of substantial noise on the learning targets projected $h$ steps ahead. We aim at demonstrating that our proposed method effectively addresses this problem by introducing self-consistency through the concurrent optimization of errors for future horizons.


\subsection{Static evaluation: R2 on fixed datasets}

For each horizon $j$, the error is computed by considering all the sub-trajectories of size $j$ from the Test dataset. The predictions are then computed by $j$ model compositions (using the groundtruth actions) and the average R2 score is reported.

\begin{figure}[htp]
\centering
\includegraphics[width=.325\textwidth]{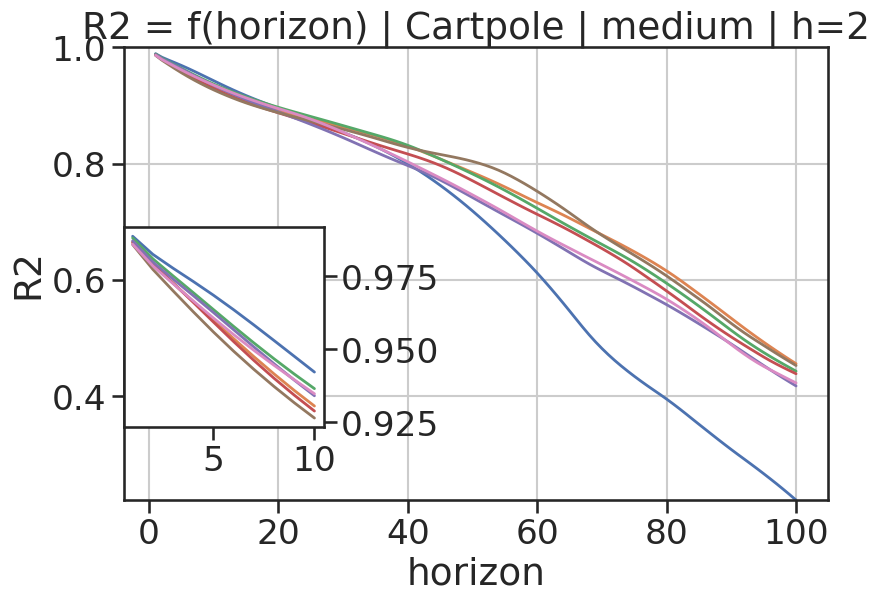}
\includegraphics[width=.325\textwidth]{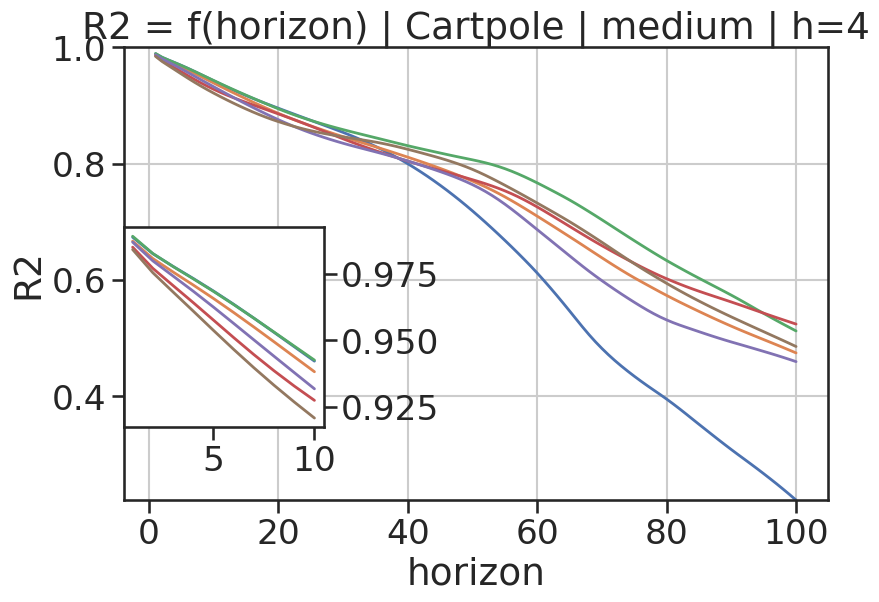}
\includegraphics[width=.325\textwidth]{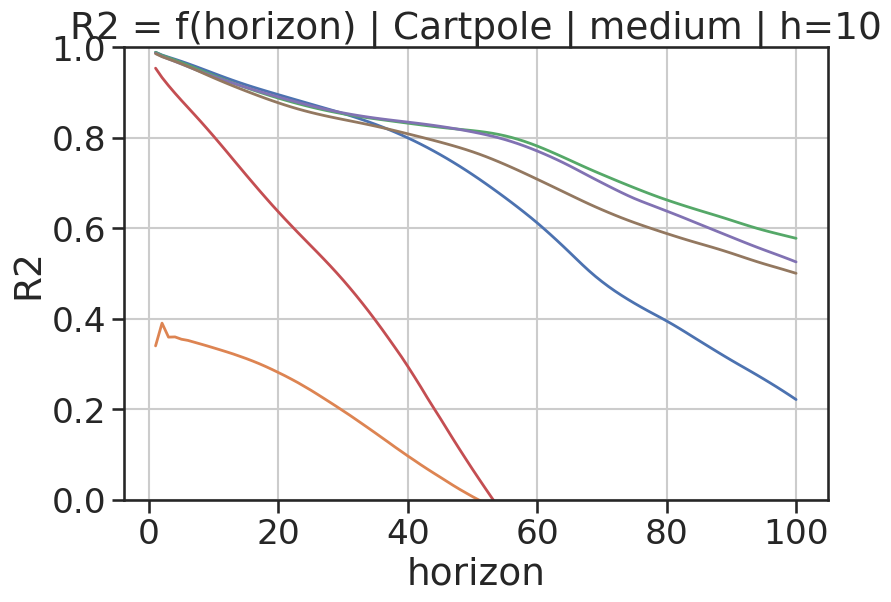}
\includegraphics[width=.325\textwidth]{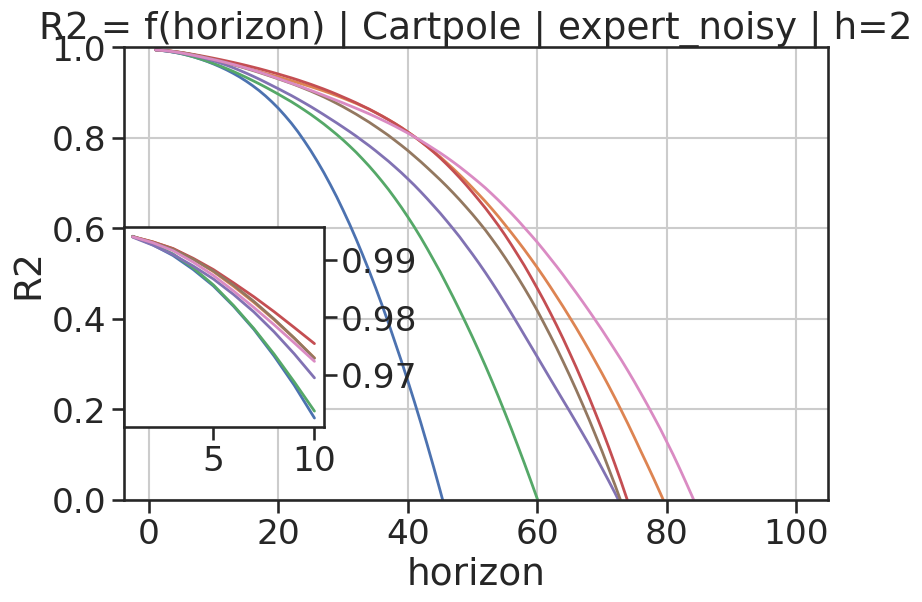}
\includegraphics[width=.325\textwidth]{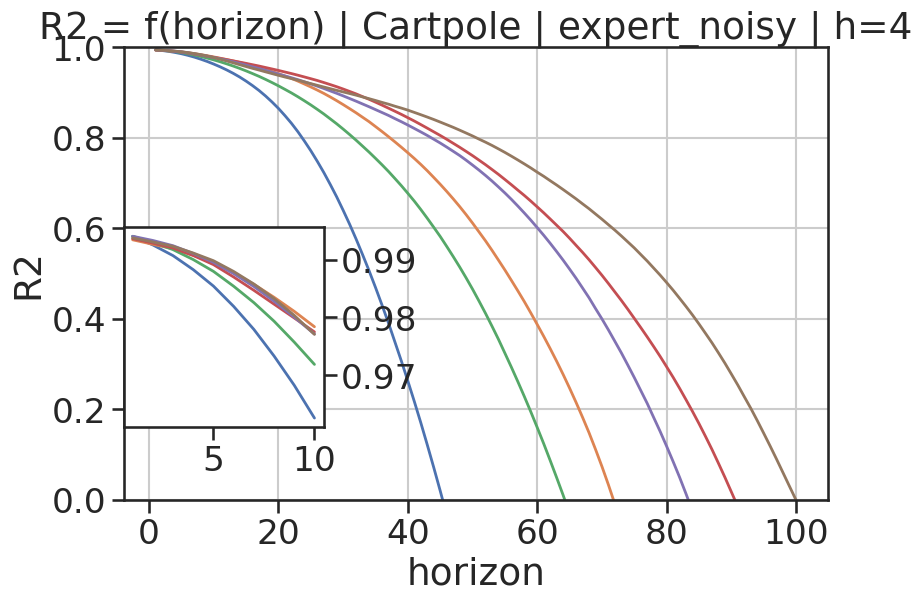}
\includegraphics[width=.325\textwidth]{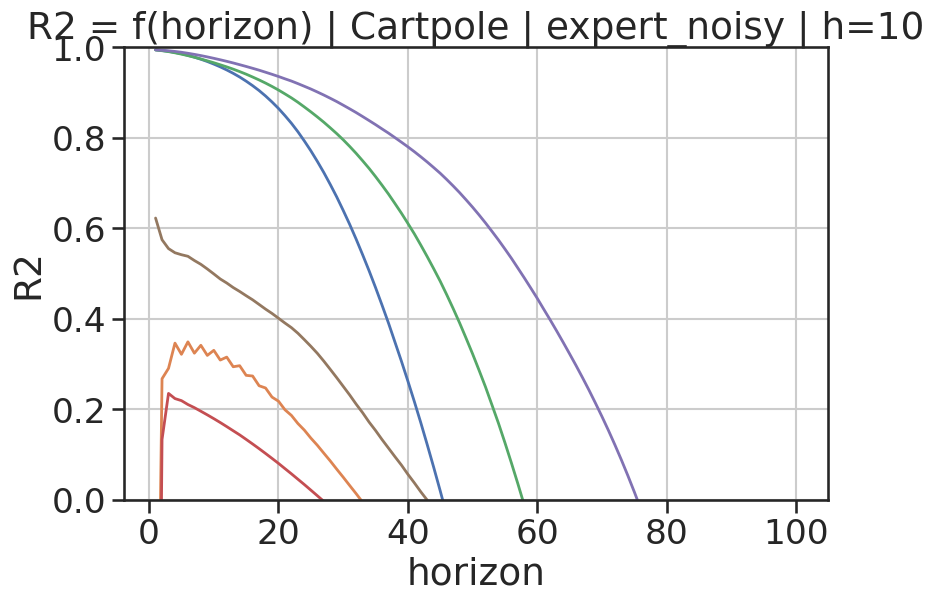}
\includegraphics[width=.8\textwidth]{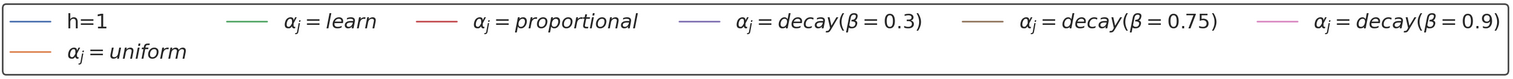}
\caption{For different maximum horizons $h \in \{2,4,10\}$ represented by columns, and datasets (medium and expert\_noisy) represented by rows, we show the R2 score as a function of the prediction horizon for different weight profiles.}
\label{fig:alpha}
\end{figure}

Figure \ref{fig:alpha} shows the R2 score as a function of the prediction horizon, when using multiple weight profiles for three maximum horizons models: $h \in \{2,4,10\}$. In both the medium and expert\_noisy datasets, the multi-timestep models largely improve over the vanilla one-step model. 
Depending on the future horizons scope, we obtain different optimal weight profiles (mostly from the noisy dataset as in the noiseless one the differences are minor): $decay(\beta=0.9)$ for $h=2$, $decay(\beta=0.75)$ for $h=4$, and $decay(\beta=0.3)$ for $h=10$. 
We interpret the effectiveness of the $decay$ weighting schema from the error profile witnessed for the one-step model. 
Indeed, the error for this model grows exponentially with the prediction horizon (a theoretical justification can be found in Theorem 1 of \cite{Venkatraman_2015}), suggesting that exponentially decaying weights (modulo the right parameter $\beta$) can correctly balance the training objective.

Inspired by this finding, we investigated the direct use of weights that are the exact inverse of the loss term they correspond to ($\alpha_j=proportional$). 
However, this technique didn't further improve the results. 
A potential explanation is the interdependence between the different losses as reducing one loss can indirectly affect another one.

Unsurprisingly, the learnable weighting profile quickly converges to putting most of the weight on the one-step error, this is expected because the one-step loss is the smallest among the other terms. 
A potential fix to this phenomena is to add a regularization term that enforces the weights to be more equally distributed. 
The elaboration of this idea is beyond the scope of this work.

\subsection{Dynamic evaluation: pure batch (offline) RL}

To demonstrate the effectiveness of multi-timestep models in practical RL scenarios, we start with the pure batch RL setting. Using the medium, random, and \textbf{expert} datasets for the \textit{Cartpole swing-up} environment, and the expert\_noisy dataset for its noisy variant. We train SAC agents on models that are themselves trained on these respective datasets, and evaluate them on the real system. Regarding the multi-timestep models, we choose the best model on each maximum horizon ($decay(\beta=0.9)$ for $h=2$, $decay(\beta=0.75)$ for $h=4$, and $decay(\beta=0.3)$ for $h=10$), and the vanilla one-step model for comparison.

\begin{table}[!ht]
  \scriptsize
  \caption{Pure batch (offline) RL evaluation: mean $\pm$ $90\%$ Gaussian confidence interval over 3 seeds. The reported score is the episodic return after training SAC on the pre-trained models for $500k$ steps}
  \label{tabDynamic}
  \centering
  \begin{tabular}{lllll}
    \toprule
    Model
    & medium
    & expert\_noisy
    & random
    & expert
    \\
    \cmidrule(r){1-5}
    $h=1$
    & 813.4 $\pm$ 19.8 
    & 580.3 $\pm$ 56.7
    & 308.7 $\pm$ 20.4
    & 807.3 $\pm$ 57.1
    \\
    $h=2|\alpha=decay(\beta=0.9)$
    & 812.8 $\pm$ 32.2  
    & 538.8 $\pm$ 75.8
    & 214.5 $\pm$ 71.1
    & 830.6 $\pm$ 44.2
    \\
    $h=4|\alpha=decay(\beta=0.75)$
    & 832.0 $\pm$ 35.7   
    & 639.09 $\pm$ 22.5
    & 178.9 $\pm$ 29.2
    & 822.1 $\pm$ 56.6
    \\
    $h=10|\alpha=decay(\beta=0.3)$
    & 842.4 $\pm$ 36.2 
    & 557.7 $\pm$ 141.2
    & 208.3 $\pm$ 42.4
    & 730.2 $\pm$ 147.8
    \\
    \bottomrule
  \end{tabular}
\end{table}

Although the results are not yet statistically significant due to a limited number of seeds, multi-timestep models improve over the one-step model in the datasets that require long-horizon accuracy (all but the random dataset where the policy keeps navigating the same region around the initial states). 

\begin{figure}[htp]
\centering
\includegraphics[width=.95\textwidth]{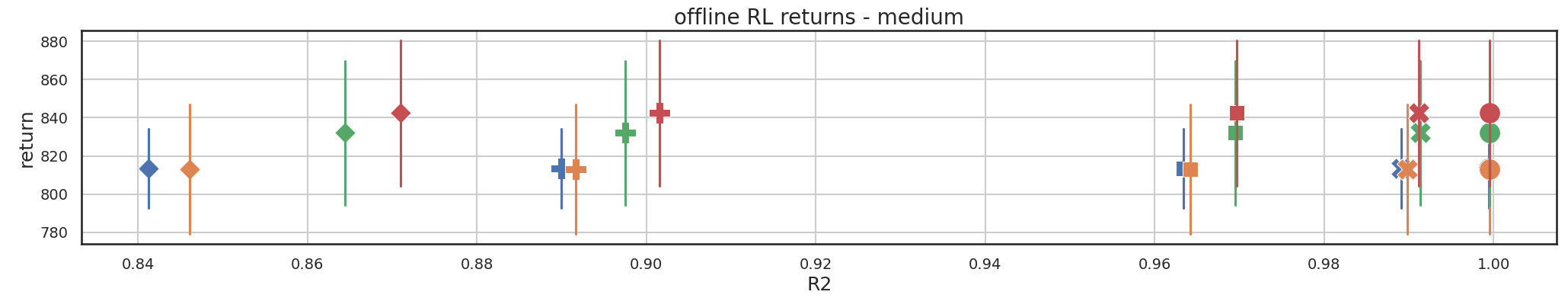}
\includegraphics[width=.95\textwidth]{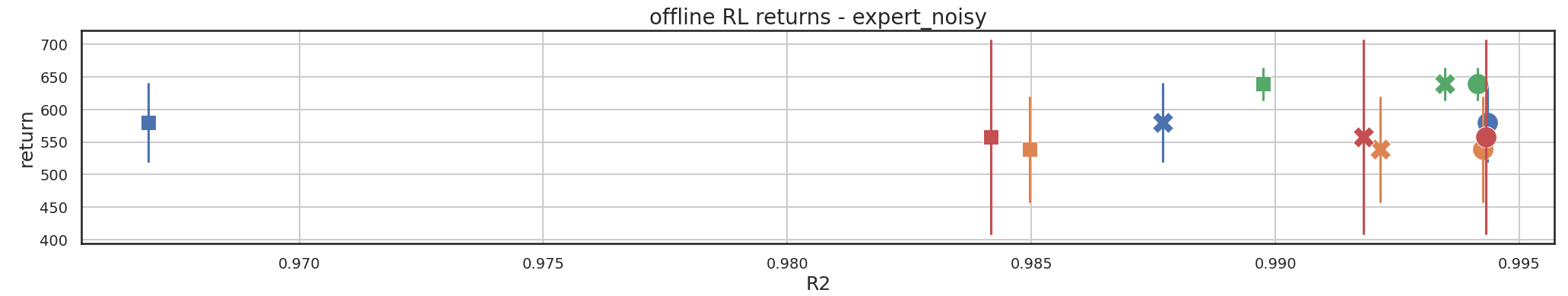}
\includegraphics[width=.8\textwidth]{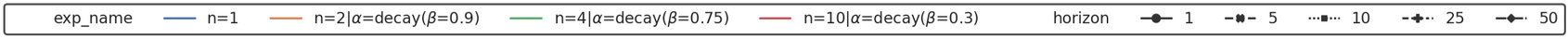}
\caption{The average episodic return as a function of the R2 score at different horizons $h \in \{1,5,10,25,50\}$. We show this for both the medium dataset representing the vanilla Cartpole environment, and expert\_noisy for its noisy variant. We omit the longer horizons ($25$ and $50$) in the second plot as the corresponding R2 scores become out of scale.}
\label{fig:offline}
\end{figure}

To get insights about the relationship between the average episodic returns of the final agents, and the predictive error of the underlying models, we propose the experiment in Figure \ref{fig:offline}. For all prediction horizons $h \in \{1,5,10,25,50\}$, we can see that in the medium dataset, the $R2(h)$ scores of the different models are positively correlated with their corresponding return. However, in the noisy dataset the relationship is not obvious. The interpretation of this empirical finding is not trivial as many factors are involved. However, a potential explanation is that related to the noise level of the environments. Indeed, models trained on noisy datasets may overfit the noise leading to worse generalizability, and consequently, worse agents in terms of the episodic return.


\subsection{Dynamic evaluation: Iterated batch RL}

\begin{figure}[htp]
\centering
\includegraphics[width=.475\textwidth]{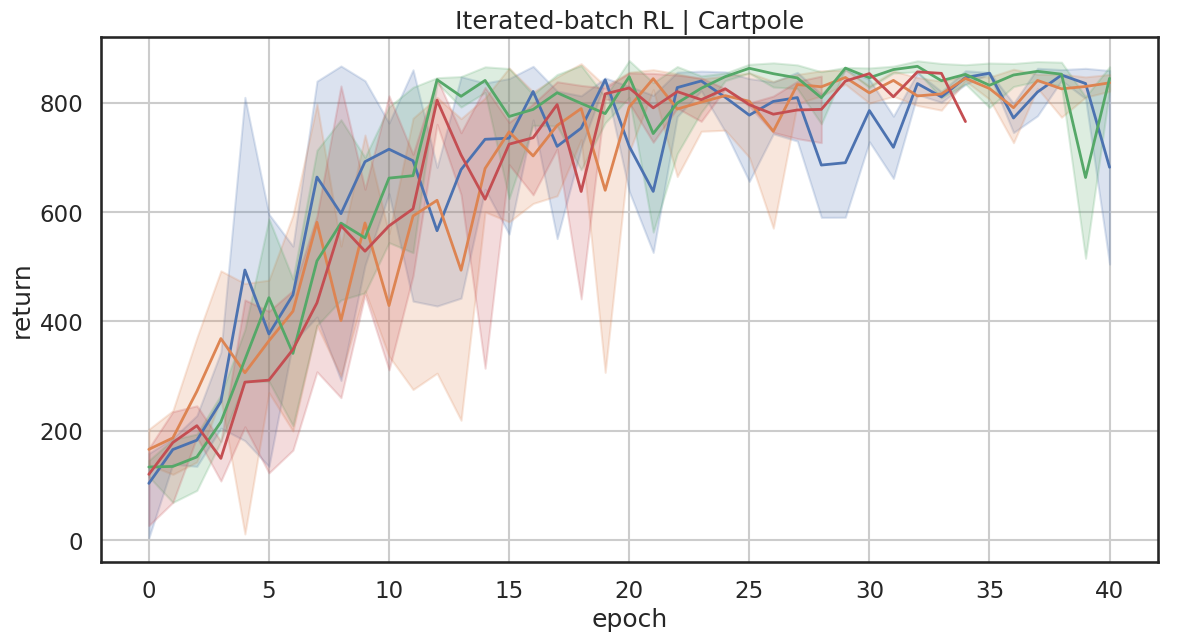}
\includegraphics[width=.475\textwidth]{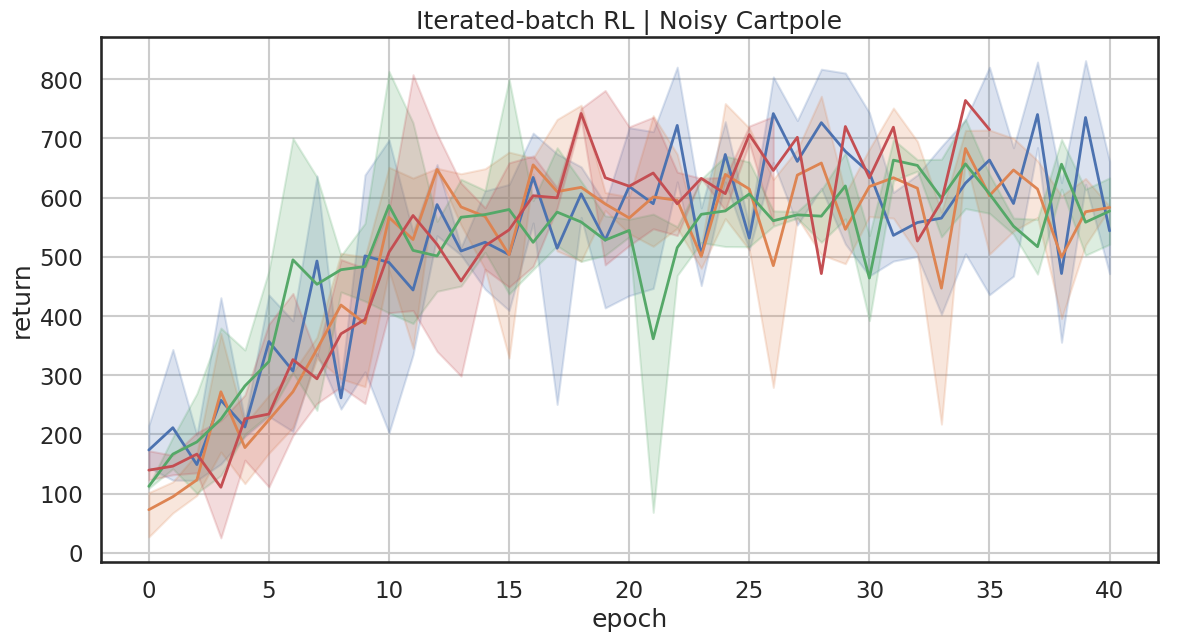}
\includegraphics[width=.8\textwidth]{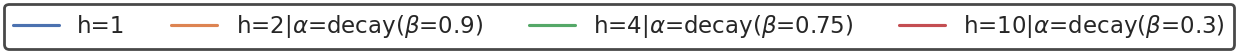}
\caption{The iterated-batch RL results for the different multi-timestep models in comparison with the standard one-step model. We report the mean $\pm$ standard deviation across 3 seeds. The \emph{epoch} in the x-axis denotes one iteration of the MBRL loop: model training on the dataset collected so far + Agent learning on model rollouts + evaluating the agent in the real system and updating the dataset.}
\label{fig:iterated}
\end{figure}

Finally, we run the full iterated batch MBRL (iterating between model learning, agent learning, and data collection) with the best multi-timestep models and compare the results with the vanilla one-step models in Figure \ref{fig:iterated}. 
In this setting, the improvement brought by the  multi-timestep models is not significant as the episodic return show a large variance among all the models.

\section{Conclusion}
\label{sesConclusion}

In this paper, we ask how to adjust the training objective of one-step predictive models to be good at generating long rollouts during planning in model-based reinforcement learning.
We build on the observation that fixed-horizon models achieve a lower predictive error down-the-horizon, and adapt the idea in the form of a novel training objective for one-step models. 
Our idea, which materialises in a weighted sum of the loss functions at future prediction horizons, led to models that, when evaluated against noiseless and noisy datasets from the Cartpole benchmark, show a large improvement of the R2 score compared to the vanilla one-step model. 
This improvement translates into better agents in pure batch (offline) RL. While the method did not improve the performance in the iterated batch setup, it did not lower the performance either. We will analyze in future work what confounding factors did not make performances improve.

One of the main research directions that follow-up from this work is the exploitation of gradient approximation schemes in the case of deterministic models. Indeed, as demonstrated in appendix \ref{AppSecGrad}, the analytical form of our proposed loss function simplifies nicely as a linear function of the standard one-step loss. Furthermore, we plan to continue evaluating the resulting multi-timestep models in different settings/environments, and test their integration with state-of-the-art MBRL algorithms that rely on single-step dynamics models.

\subsection*{Limitations}

There are two major limitations of this study that could be addressed in future research. First, the study focused solely on the Cartpole benchmark, a relatively low-dimensional environment that reduces the modelling complexity that may be encountered in other high-dimensional systems. Nevertheless, to improve the generalisability of our results, we used different offline datasets and considered a more challenging Cartpole variant with additive Gaussian noise. Second, since our proposed objective can be plugged into any single-step model-based algorithm, we believe it would be interesting to test its integration with some of the state-of-the-art MBRL algorithms in different setups/applications. This is particularly important as our experiments in the offline and iterated-batch settings did not show a statistically significant improvement. Nevertheless, in this work we showcased the importance of multi-timestep models for the dynamics modeling task using different challenging setups, suggesting that they may well find RL applications, where the dynamics modeling is crucial for the final task.

\newpage

\section*{Reproducibility Statement}
In order to ensure reproducibility we will release the code at \texttt{<URL hidden for review>}, once the paper has been accepted. The implementation details and hyperparameters are listed in Appendix \ref{appendixsecImpDetails}.

\bibliography{main}

\begin{thebibliography}{55}
\providecommand{\natexlab}[1]{#1}
\providecommand{\url}[1]{\texttt{#1}}
\expandafter\ifx\csname urlstyle\endcsname\relax
  \providecommand{\doi}[1]{doi: #1}\else
  \providecommand{\doi}{doi: \begingroup \urlstyle{rm}\Url}\fi

\bibitem[Argenson \& Dulac-Arnold(2021)Argenson and Dulac-Arnold]{Argenson2021}
Arthur Argenson and Gabriel Dulac-Arnold.
\newblock Model-based offline planning.
\newblock In \emph{International Conference on Learning Representations}, 2021.

\bibitem[Asadi et~al.(2018)Asadi, Cater, Misra, and Littman]{Asadi_2018}
Kavosh Asadi, Evan Cater, Dipendra Misra, and Michael~L. Littman.
\newblock Towards a {Simple} {Approach} to {Multi}-step {Model}-based {Reinforcement} {Learning}, October 2018.
\newblock URL \url{http://arxiv.org/abs/1811.00128}.
\newblock arXiv:1811.00128 [cs, stat].

\bibitem[Asadi et~al.(2019)Asadi, Misra, Kim, and Littman]{Asadi_2019}
Kavosh Asadi, Dipendra Misra, Seungchan Kim, and Michel~L. Littman.
\newblock Combating the {Compounding}-{Error} {Problem} with a {Multi}-step {Model}, May 2019.
\newblock URL \url{http://arxiv.org/abs/1905.13320}.
\newblock arXiv:1905.13320 [cs, stat].

\bibitem[Ben~Taieb \& Bontempi(2012)Ben~Taieb and Bontempi]{Bentaieb_2012a}
Souhaib Ben~Taieb and Gianluca Bontempi.
\newblock Recursive {Multi}-step {Time} {Series} {Forecasting} by {Perturbing} {Data}, January 2012.
\newblock URL \url{https://ieeexplore.ieee.org/abstract/document/6137274}.

\bibitem[Ben~Taieb et~al.(2012)Ben~Taieb, Bontempi, Atiya, and Sorjamaa]{Bentaieb_2012b}
Souhaib Ben~Taieb, Gianluca Bontempi, Amir~F. Atiya, and Antti Sorjamaa.
\newblock A review and comparison of strategies for multi-step ahead time series forecasting based on the {NN5} forecasting competition.
\newblock \emph{Expert Systems with Applications}, 39\penalty0 (8):\penalty0 7067--7083, June 2012.
\newblock ISSN 0957-4174.
\newblock \doi{10.1016/j.eswa.2012.01.039}.
\newblock URL \url{https://www.sciencedirect.com/science/article/pii/S0957417412000528}.

\bibitem[Benechehab et~al.(2023)Benechehab, Thomas, and Kégl]{Benechehab_2023}
Abdelhakim Benechehab, Albert Thomas, and Balázs Kégl.
\newblock Deep autoregressive density nets vs neural ensembles for model-based offline reinforcement learning.
\newblock February 2023.
\newblock URL \url{https://openreview.net/forum?id=gvOSQjGTtxj}.

\bibitem[Byravan et~al.(2021)Byravan, Hasenclever, Trochim, Mirza, Ialongo, Tassa, Springenberg, Abdolmaleki, Heess, Merel, and Riedmiller]{Byravan_2021}
Arunkumar Byravan, Leonard Hasenclever, Piotr Trochim, Mehdi Mirza, Alessandro~Davide Ialongo, Yuval Tassa, Jost~Tobias Springenberg, Abbas Abdolmaleki, Nicolas Heess, Josh Merel, and Martin~A. Riedmiller.
\newblock Evaluating model-based planning and planner amortization for continuous control.
\newblock \emph{CoRR}, abs/2110.03363, 2021.
\newblock URL \url{https://arxiv.org/abs/2110.03363}.

\bibitem[Chandra et~al.(2021)Chandra, Goyal, and Gupta]{Chandra_2021}
Rohitash Chandra, Shaurya Goyal, and Rishabh Gupta.
\newblock Evaluation of deep learning models for multi-step ahead time series prediction.
\newblock \emph{IEEE Access}, 9:\penalty0 83105--83123, 2021.
\newblock ISSN 2169-3536.
\newblock \doi{10.1109/ACCESS.2021.3085085}.
\newblock URL \url{http://arxiv.org/abs/2103.14250}.
\newblock arXiv:2103.14250 [cs].

\bibitem[Chua et~al.(2018)Chua, Calandra, McAllister, and Levine]{Chua2018}
Kurtland Chua, Roberto Calandra, Rowan McAllister, and Sergey Levine.
\newblock Deep reinforcement learning in a handful of trials using probabilistic dynamics models.
\newblock In \emph{Advances in Neural Information Processing Systems 31}, pp.\  4754--4765. Curran Associates, Inc., 2018.

\bibitem[Chung et~al.(2014)Chung, G{\"{u}}l{\c{c}}ehre, Cho, and Bengio]{Chung_2014}
Junyoung Chung, {\c{C}}aglar G{\"{u}}l{\c{c}}ehre, KyungHyun Cho, and Yoshua Bengio.
\newblock Empirical evaluation of gated recurrent neural networks on sequence modeling.
\newblock \emph{CoRR}, abs/1412.3555, 2014.
\newblock URL \url{http://arxiv.org/abs/1412.3555}.

\bibitem[Deisenroth \& Rasmussen(2011)Deisenroth and Rasmussen]{Deisenroth2011}
Marc~Peter Deisenroth and Carl~Edward Rasmussen.
\newblock {PILCO}: A model-based and data-efficient approach to policy search.
\newblock In \emph{Proceedings of the International Conference on Machine Learning}, 2011.

\bibitem[Draeger et~al.(1995)Draeger, Engell, and Ranke]{Draeger1995}
Andreas Draeger, Sebastian Engell, and Horst Ranke.
\newblock Model predictive control using neural networks.
\newblock \emph{IEEE Control Systems}, 15:\penalty0 61--66, 1995.
\newblock ISSN 1066033X.
\newblock \doi{10.1109/37.466261}.

\bibitem[Gal et~al.(2016)Gal, McAllister, and Rasmussen]{Gal2016}
Yarin Gal, Rowan McAllister, and Carl~Edward Rasmussen.
\newblock Improving {PILCO} with {B}ayesian neural network dynamics models.
\newblock In \emph{Data-Efficient Machine Learning workshop, International Conference on Machine Learning}, 2016.

\bibitem[Ha \& Schmidhuber(2018)Ha and Schmidhuber]{Ha2018}
David Ha and J\"{u}rgen Schmidhuber.
\newblock Recurrent world models facilitate policy evolution.
\newblock In S.~Bengio, H.~Wallach, H.~Larochelle, K.~Grauman, N.~Cesa-Bianchi, and R.~Garnett (eds.), \emph{Advances in Neural Information Processing Systems 31}, pp.\  2450--2462. Curran Associates, Inc., 2018.

\bibitem[Hafner et~al.(2019)Hafner, Lillicrap, Fischer, Villegas, Ha, Lee, and Davidson]{HafneR2019}
Danijar Hafner, Timothy Lillicrap, Ian Fischer, Ruben Villegas, David Ha, Honglak Lee, and James Davidson.
\newblock Learning latent dynamics for planning from pixels.
\newblock In \emph{Proceedings of the 36th International Conference on Machine Learning}, volume~97 of \emph{Proceedings of Machine Learning Research}, pp.\  2555--2565, 2019.

\bibitem[Hafner et~al.(2021)Hafner, Lillicrap, Norouzi, and Ba]{HafneR2021}
Danijar Hafner, Timothy~P Lillicrap, Mohammad Norouzi, and Jimmy Ba.
\newblock Mastering atari with discrete world models.
\newblock In \emph{International Conference on Learning Representations}, 2021.
\newblock URL \url{https://openreview.net/forum?id=0oabwyZbOu}.

\bibitem[Hochreiter \& Schmidhuber(1997)Hochreiter and Schmidhuber]{Hochreiter_1997}
Sepp Hochreiter and Jürgen Schmidhuber.
\newblock Long short-term memory.
\newblock \emph{Neural Computation}, 9\penalty0 (8):\penalty0 1735--1780, 1997.

\bibitem[Ioffe \& Szegedy(2015)Ioffe and Szegedy]{Ioffe2015}
Sergey Ioffe and Christian Szegedy.
\newblock Batch normalization: Accelerating deep network training by reducing internal covariate shift.
\newblock In \emph{Proceedings of the 32nd International Conference on International Conference on Machine Learning - Volume 37}, ICML'15, pp.\  448–456. JMLR.org, 2015.

\bibitem[Janner et~al.(2019)Janner, Fu, Zhang, and Levine]{JanneR2019}
Michael Janner, Justin Fu, Marvin Zhang, and Sergey Levine.
\newblock When to trust your model: Model-based policy optimization.
\newblock In H.~Wallach, H.~Larochelle, A.~Beygelzimer, F.~d\textquotesingle Alch\'{e}-Buc, E.~Fox, and R.~Garnett (eds.), \emph{Advances in Neural Information Processing Systems}, volume~32. Curran Associates, Inc., 2019.

\bibitem[K{\'e}gl et~al.(2018)K{\'e}gl, Boucaud, Cherti, Kazakci, Gramfort, Lemaitre, den Bossche, Benbouzid, and Marini]{Kegl2018}
Bal{\'a}zs K{\'e}gl, Alexandre Boucaud, Mehdi Cherti, Akin Kazakci, Alexandre Gramfort, Guillaume Lemaitre, Joris~Van den Bossche, Djalel Benbouzid, and Camille Marini.
\newblock The {RAMP} framework: from reproducibility to transparency in the design and optimization of scientific workflows.
\newblock In \emph{ICML workshop on Reproducibility in Machine Learning}, 2018.

\bibitem[K{\'e}gl et~al.(2021)K{\'e}gl, Hurtado, and Thomas]{Kegl2021}
Bal{\'a}zs K{\'e}gl, Gabriel Hurtado, and Albert Thomas.
\newblock Model-based micro-data reinforcement learning: what are the crucial model properties and which model to choose?
\newblock In \emph{International Conference on Learning Representations}, 2021.
\newblock URL \url{https://openreview.net/forum?id=p5uylG94S68}.

\bibitem[Kidambi et~al.(2020)Kidambi, Rajeswaran, Netrapalli, and Joachims]{Kidambi2020}
Rahul Kidambi, Aravind Rajeswaran, Praneeth Netrapalli, and Thorsten Joachims.
\newblock Morel: Model-based offline reinforcement learning.
\newblock In H.~Larochelle, M.~Ranzato, R.~Hadsell, M.F. Balcan, and H.~Lin (eds.), \emph{Advances in Neural Information Processing Systems}, volume~33, pp.\  21810--21823. Curran Associates, Inc., 2020.
\newblock URL \url{https://proceedings.neurips.cc/paper/2020/file/f7efa4f864ae9b88d43527f4b14f750f-Paper.pdf}.

\bibitem[Kingma \& Ba(2015)Kingma and Ba]{Diederik2015}
Diederik~P. Kingma and Jimmy Ba.
\newblock Adam: {A} method for stochastic optimization.
\newblock In Yoshua Bengio and Yann LeCun (eds.), \emph{3rd International Conference on Learning Representations, {ICLR} 2015, San Diego, CA, USA, May 7-9, 2015, Conference Track Proceedings}, 2015.
\newblock URL \url{http://arxiv.org/abs/1412.6980}.

\bibitem[Lambert et~al.(2021)Lambert, Wilcox, Zhang, Pister, and Calandra]{Lambert_2021}
Nathan Lambert, Albert Wilcox, Howard Zhang, Kristofer S.~J. Pister, and Roberto Calandra.
\newblock Learning {Accurate} {Long}-term {Dynamics} for {Model}-based {Reinforcement} {Learning}.
\newblock In \emph{2021 60th {IEEE} {Conference} on {Decision} and {Control} ({CDC})}, pp.\  2880--2887, December 2021.
\newblock \doi{10.1109/CDC45484.2021.9683134}.
\newblock ISSN: 2576-2370.

\bibitem[Lambert et~al.(2022)Lambert, Pister, and Calandra]{Lambert_2022}
Nathan Lambert, Kristofer Pister, and Roberto Calandra.
\newblock Investigating {Compounding} {Prediction} {Errors} in {Learned} {Dynamics} {Models}, March 2022.
\newblock URL \url{http://arxiv.org/abs/2203.09637}.
\newblock arXiv:2203.09637 [cs].

\bibitem[Lee et~al.(2021)Lee, Lee, and Kim]{Lee2021}
Byung-Jun Lee, Jongmin Lee, and Kee-Eung Kim.
\newblock Representation balancing offline model-based reinforcement learning.
\newblock In \emph{International Conference on Learning Representations}, 2021.
\newblock URL \url{https://openreview.net/forum?id=QpNz8r_Ri2Y}.

\bibitem[Levine \& Koltun(2013)Levine and Koltun]{Levine2013}
Sergey Levine and Vladlen Koltun.
\newblock Guided policy search.
\newblock In Sanjoy Dasgupta and David McAllester (eds.), \emph{Proceedings of the 30th International Conference on Machine Learning}, volume~28 of \emph{Proceedings of Machine Learning Research}, pp.\  1--9, Atlanta, Georgia, USA, 17--19 Jun 2013. PMLR.
\newblock URL \url{https://proceedings.mlr.press/v28/levine13.html}.

\bibitem[Liu et~al.(2021)Liu, Chen, and Zhong]{Liu2021}
Ruizhen Liu, Zhicong Chen, and Dazhi Zhong.
\newblock Dromo: Distributionally robust offline model-based policy optimization.
\newblock 2021.

\bibitem[Mnih et~al.(2015)Mnih, Kavukcuoglu, Silver, Rusu, Veness, Bellemare, Graves, Riedmiller, Fidjeland, Ostrovski, Petersen, Beattie, Sadik, Antonoglou, King, Kumaran, Wierstra, Legg, and Hassabis]{Mnih2015}
Volodymyr Mnih, Koray Kavukcuoglu, David Silver, Andrei~A. Rusu, Joel Veness, Marc~G. Bellemare, Alex Graves, Martin Riedmiller, Andreas~K. Fidjeland, Georg Ostrovski, Stig Petersen, Charles Beattie, Amir Sadik, Ioannis Antonoglou, Helen King, Dharshan Kumaran, Daan Wierstra, Shane Legg, and Demis Hassabis.
\newblock Human-level control through deep reinforcement learning.
\newblock \emph{Nature}, 518\penalty0 (7540):\penalty0 529--533, 2015.

\bibitem[Nagabandi et~al.(2018)Nagabandi, Kahn, Fearing, and Levine]{Nagabandi2018}
Anusha Nagabandi, Gregory Kahn, Ronald~S. Fearing, and Sergey Levine.
\newblock Neural network dynamics for model-based deep reinforcement learning with model-free fine-tuning.
\newblock In \emph{2018 {IEEE} International Conference on Robotics and Automation, {ICRA} 2018}, pp.\  7559--7566. {IEEE}, 2018.

\bibitem[Pinneri et~al.(2020)Pinneri, Sawant, Blaes, Achterhold, Stueckler, Rolinek, and Martius]{Pinneri2020}
Cristina Pinneri, Shambhuraj Sawant, Sebastian Blaes, Jan Achterhold, Joerg Stueckler, Michal Rolinek, and Georg Martius.
\newblock Sample-efficient cross-entropy method for real-time planning.
\newblock In \emph{Conference on Robot Learning 2020}, 2020.
\newblock URL \url{https://corlconf.github.io/corl2020/paper_217/}.

\bibitem[Precup \& Sutton(1997)Precup and Sutton]{Precup1997}
Doina Precup and Richard~S Sutton.
\newblock Multi-time {Models} for {Temporally} {Abstract} {Planning}.
\newblock In \emph{Advances in {Neural} {Information} {Processing} {Systems}}, volume~10. MIT Press, 1997.
\newblock URL \url{https://proceedings.neurips.cc/paper_files/paper/1997/hash/a9be4c2a4041cadbf9d61ae16dd1389e-Abstract.html}.

\bibitem[Precup et~al.(1998)Precup, Sutton, and Singh]{Precup1998}
Doina Precup, Richard~S. Sutton, and Satinder Singh.
\newblock Theoretical results on reinforcement learning with temporally abstract options.
\newblock In Claire Nédellec and Céline Rouveirol (eds.), \emph{Machine {Learning}: {ECML}-98}, Lecture {Notes} in {Computer} {Science}, pp.\  382--393, Berlin, Heidelberg, 1998. Springer.
\newblock ISBN 978-3-540-69781-7.
\newblock \doi{10.1007/BFb0026709}.

\bibitem[Raffin et~al.(2021)Raffin, Hill, Gleave, Kanervisto, Ernestus, and Dormann]{Raffin2021}
Antonin Raffin, Ashley Hill, Adam Gleave, Anssi Kanervisto, Maximilian Ernestus, and Noah Dormann.
\newblock Stable-baselines3: Reliable reinforcement learning implementations.
\newblock \emph{Journal of Machine Learning Research}, 22\penalty0 (268):\penalty0 1--8, 2021.
\newblock URL \url{http://jmlr.org/papers/v22/20-1364.html}.

\bibitem[Silver et~al.(2017{\natexlab{a}})Silver, Hasselt, Hessel, Schaul, Guez, Harley, Dulac-Arnold, Reichert, Rabinowitz, Barreto, and Degris]{Silver_2017}
David Silver, Hado Hasselt, Matteo Hessel, Tom Schaul, Arthur Guez, Tim Harley, Gabriel Dulac-Arnold, David Reichert, Neil Rabinowitz, Andre Barreto, and Thomas Degris.
\newblock The {Predictron}: {End}-{To}-{End} {Learning} and {Planning}.
\newblock In \emph{Proceedings of the 34th {International} {Conference} on {Machine} {Learning}}, pp.\  3191--3199. PMLR, July 2017{\natexlab{a}}.
\newblock URL \url{https://proceedings.mlr.press/v70/silver17a.html}.
\newblock ISSN: 2640-3498.

\bibitem[Silver et~al.(2017{\natexlab{b}})Silver, Hubert, Schrittwieser, Antonoglou, Lai, Guez, Lanctot, Sifre, Kumaran, Graepel, Lillicrap, Simonyan, and Hassabis]{SilveR2017}
David Silver, Thomas Hubert, Julian Schrittwieser, Ioannis Antonoglou, Matthew Lai, Arthur Guez, Marc Lanctot, Laurent Sifre, Dharshan Kumaran, Thore Graepel, Timothy Lillicrap, Karen Simonyan, and Demis Hassabis.
\newblock Mastering chess and shogi by self-play with a general reinforcement learning algorithm, 2017{\natexlab{b}}.

\bibitem[Silver et~al.(2018)Silver, Hubert, Schrittwieser, Antonoglou, Lai, Guez, Lanctot, Sifre, Kumaran, Graepel, Lillicrap, Simonyan, and Hassabis]{SilveR2018}
David Silver, Thomas Hubert, Julian Schrittwieser, Ioannis Antonoglou, Matthew Lai, Arthur Guez, Marc Lanctot, Laurent Sifre, Dharshan Kumaran, Thore Graepel, Timothy Lillicrap, Karen Simonyan, and Demis Hassabis.
\newblock A general reinforcement learning algorithm that masters {C}hess, {S}hogi, and {Go} through self-play.
\newblock \emph{Science}, 362\penalty0 (6419):\penalty0 1140--1144, 2018.
\newblock ISSN 0036-8075.
\newblock \doi{10.1126/science.aar6404}.

\bibitem[Singh(1992)]{Singh1992}
Satinder~P. Singh.
\newblock Scaling {Reinforcement} {Learning} {Algorithms} by {Learning} {Variable} {Temporal} {Resolution} {Models}.
\newblock In Derek Sleeman and Peter Edwards (eds.), \emph{Machine {Learning} {Proceedings} 1992}, pp.\  406--415. Morgan Kaufmann, San Francisco (CA), January 1992.
\newblock ISBN 978-1-55860-247-2.
\newblock \doi{10.1016/B978-1-55860-247-2.50058-9}.
\newblock URL \url{https://www.sciencedirect.com/science/article/pii/B9781558602472500589}.

\bibitem[Srivastava et~al.(2014)Srivastava, Hinton, Krizhevsky, Sutskever, and Salakhutdinov]{Srivastava2014}
Nitish Srivastava, Geoffrey Hinton, Alex Krizhevsky, Ilya Sutskever, and Ruslan Salakhutdinov.
\newblock Dropout: A simple way to prevent neural networks from overfitting.
\newblock \emph{Journal of Machine Learning Research}, 15\penalty0 (56):\penalty0 1929--1958, 2014.
\newblock URL \url{http://jmlr.org/papers/v15/srivastava14a.html}.

\bibitem[Sutton(1991)]{Sutton1991}
Richard~S. Sutton.
\newblock Dyna, an integrated architecture for learning, planning, and reacting.
\newblock \emph{ACM SIGART Bulletin}, 2:\penalty0 160--163, 7 1991.
\newblock ISSN 0163-5719.
\newblock \doi{10.1145/122344.122377}.
\newblock URL \url{https://dl.acm.org/doi/10.1145/122344.122377}.

\bibitem[Sutton(1995)]{sutton1995}
Richard~S. Sutton.
\newblock {TD} {Models}: {Modeling} the {World} at a {Mixture} of {Time} {Scales}.
\newblock In Armand Prieditis and Stuart Russell (eds.), \emph{Machine {Learning} {Proceedings} 1995}, pp.\  531--539. Morgan Kaufmann, San Francisco (CA), January 1995.
\newblock ISBN 978-1-55860-377-6.
\newblock \doi{10.1016/B978-1-55860-377-6.50072-4}.
\newblock URL \url{https://www.sciencedirect.com/science/article/pii/B9781558603776500724}.

\bibitem[Sutton \& Pinette(1985)Sutton and Pinette]{Sutton1985}
Richard~S Sutton and Brian Pinette.
\newblock The learning of world models by connectionist networks, 1985.

\bibitem[Sutton et~al.(1992)Sutton, Szepesvári, Geramifard, and Bowling]{Sutton1992}
Richard~S Sutton, Csaba Szepesvári, Alborz Geramifard, and Michael Bowling.
\newblock Dyna-style planning with linear function approximation and prioritized sweeping.
\newblock \emph{Moore and Atkeson}, 1992.

\bibitem[Sutton et~al.(1999)Sutton, Precup, and Singh]{Sutton1999}
Richard~S. Sutton, Doina Precup, and Satinder Singh.
\newblock Between {MDPs} and semi-{MDPs}: {A} framework for temporal abstraction in reinforcement learning.
\newblock \emph{Artificial Intelligence}, 112\penalty0 (1):\penalty0 181--211, August 1999.
\newblock ISSN 0004-3702.
\newblock \doi{10.1016/S0004-3702(99)00052-1}.
\newblock URL \url{https://www.sciencedirect.com/science/article/pii/S0004370299000521}.

\bibitem[Talvitie(2014)]{Talvitie_2014}
Erik Talvitie.
\newblock Model {Regularization} for {Stable} {Sample} {Rollouts}.
\newblock 2014.

\bibitem[Talvitie(2017)]{Talvitie_2017}
Erik Talvitie.
\newblock Self-{Correcting} {Models} for {Model}-{Based} {Reinforcement} {Learning}.
\newblock \emph{Proceedings of the AAAI Conference on Artificial Intelligence}, 31\penalty0 (1), February 2017.
\newblock ISSN 2374-3468, 2159-5399.
\newblock \doi{10.1609/aaai.v31i1.10850}.
\newblock URL \url{https://ojs.aaai.org/index.php/AAAI/article/view/10850}.

\bibitem[Tassa et~al.(2018)Tassa, Doron, Muldal, Erez, Li, de~Las~Casas, Budden, Abdolmaleki, Merel, Lefrancq, Lillicrap, and Riedmiller]{Tassa_2018}
Yuval Tassa, Yotam Doron, Alistair Muldal, Tom Erez, Yazhe Li, Diego de~Las~Casas, David Budden, Abbas Abdolmaleki, Josh Merel, Andrew Lefrancq, Timothy~P. Lillicrap, and Martin~A. Riedmiller.
\newblock Deepmind control suite.
\newblock \emph{CoRR}, abs/1801.00690, 2018.
\newblock URL \url{http://arxiv.org/abs/1801.00690}.

\bibitem[Todorov et~al.(2012)Todorov, Erez, and Tassa]{Todorov2012}
Emanuel Todorov, Tom Erez, and Yuval Tassa.
\newblock Mujoco: A physics engine for model-based control.
\newblock In \emph{2012 IEEE/RSJ International Conference on Intelligent Robots and Systems}, pp.\  5026--5033, 2012.
\newblock \doi{10.1109/IROS.2012.6386109}.

\bibitem[Venkatraman et~al.(2015)Venkatraman, Hebert, and Bagnell]{Venkatraman_2015}
Arun Venkatraman, Martial Hebert, and J.. Bagnell.
\newblock Improving {Multi}-{Step} {Prediction} of {Learned} {Time} {Series} {Models}.
\newblock \emph{Proceedings of the AAAI Conference on Artificial Intelligence}, 29\penalty0 (1), February 2015.
\newblock ISSN 2374-3468, 2159-5399.
\newblock \doi{10.1609/aaai.v29i1.9590}.
\newblock URL \url{https://ojs.aaai.org/index.php/AAAI/article/view/9590}.

\bibitem[Vinyals et~al.(2019)Vinyals, Babuschkin, Czarnecki, Mathieu, Dudzik, Chung, Choi, Powell, Ewalds, Georgiev, Oh, Horgan, Kroiss, Danihelka, Huang, Sifre, Cai, Agapiou, Jaderberg, Vezhnevets, Leblond, Pohlen, Dalibard, Budden, Sulsky, Molloy, Paine, Gulcehre, Wang, Pfaff, Wu, Ring, Yogatama, W{\"u}nsch, McKinney, Smith, Schaul, Lillicrap, Kavukcuoglu, Hassabis, Apps, and Silver]{Vinyals2019}
Oriol Vinyals, Igor Babuschkin, Wojciech~M. Czarnecki, Micha{\"e}l Mathieu, Andrew Dudzik, Junyoung Chung, David~H. Choi, Richard Powell, Timo Ewalds, Petko Georgiev, Junhyuk Oh, Dan Horgan, Manuel Kroiss, Ivo Danihelka, Aja Huang, L.~Sifre, Trevor Cai, John~P. Agapiou, Max Jaderberg, Alexander~Sasha Vezhnevets, R{\'e}mi Leblond, Tobias Pohlen, Valentin Dalibard, David Budden, Yury Sulsky, James Molloy, Tom~Le Paine, Caglar Gulcehre, Ziyun Wang, Tobias Pfaff, Yuhuai Wu, Roman Ring, Dani Yogatama, Dario W{\"u}nsch, Katrina McKinney, Oliver Smith, Tom Schaul, Timothy~P. Lillicrap, Koray Kavukcuoglu, Demis Hassabis, Chris Apps, and David Silver.
\newblock Grandmaster level in starcraft ii using multi-agent reinforcement learning.
\newblock \emph{Nature}, pp.\  1--5, 2019.

\bibitem[Whitney \& Fergus(2019)Whitney and Fergus]{Whitney_2019}
William Whitney and Rob Fergus.
\newblock Understanding the {Asymptotic} {Performance} of {Model}-{Based} {RL} {Methods}.
\newblock May 2019.
\newblock URL \url{https://openreview.net/forum?id=B1g29oAqtm}.

\bibitem[Xiao et~al.(2023)Xiao, Wu, Ma, Schuurmans, and Müller]{Xiao_2023}
Chenjun Xiao, Yifan Wu, Chen Ma, Dale Schuurmans, and Martin Müller.
\newblock Learning to {Combat} {Compounding}-{Error} in {Model}-{Based} {Reinforcement} {Learning}.
\newblock May 2023.
\newblock URL \url{https://openreview.net/forum?id=S1g_S0VYvr}.

\bibitem[Yu et~al.(2020)Yu, Thomas, Yu, Ermon, Zou, Levine, Finn, and Ma]{Yu2020}
Tianhe Yu, Garrett Thomas, Lantao Yu, Stefano Ermon, James~Y Zou, Sergey Levine, Chelsea Finn, and Tengyu Ma.
\newblock Mopo: Model-based offline policy optimization.
\newblock In H.~Larochelle, M.~Ranzato, R.~Hadsell, M.F. Balcan, and H.~Lin (eds.), \emph{Advances in Neural Information Processing Systems}, volume~33, pp.\  14129--14142. Curran Associates, Inc., 2020.
\newblock URL \url{https://proceedings.neurips.cc/paper/2020/file/a322852ce0df73e204b7e67cbbef0d0a-Paper.pdf}.

\bibitem[Yu et~al.(2021)Yu, Kumar, Rafailov, Rajeswaran, Levine, and Finn]{Yu2021}
Tianhe Yu, Aviral Kumar, Rafael Rafailov, Aravind Rajeswaran, Sergey Levine, and Chelsea Finn.
\newblock Combo: Conservative offline model-based policy optimization.
\newblock In M.~Ranzato, A.~Beygelzimer, Y.~Dauphin, P.S. Liang, and J.~Wortman Vaughan (eds.), \emph{Advances in Neural Information Processing Systems}, volume~34, pp.\  28954--28967. Curran Associates, Inc., 2021.
\newblock URL \url{https://proceedings.neurips.cc/paper/2021/file/f29a179746902e331572c483c45e5086-Paper.pdf}.

\bibitem[Zhan et~al.(2021)Zhan, Zhu, and Xu]{Zhan2021}
Xianyuan Zhan, Xiangyu Zhu, and Haoran Xu.
\newblock Model-based offline planning with trajectory pruning.
\newblock 2021.

\end{thebibliography}
\bibliographystyle{main}

\newpage

\appendix

\section{The Cartpole Reinforcement Learning Benchmark}
\label{appendixcartpole}

\subsection{The environment}

\begin{wrapfigure}{r}{0.25\textwidth}
\vspace{-5mm}
  \begin{center}
    \includegraphics[width=0.2\textwidth]{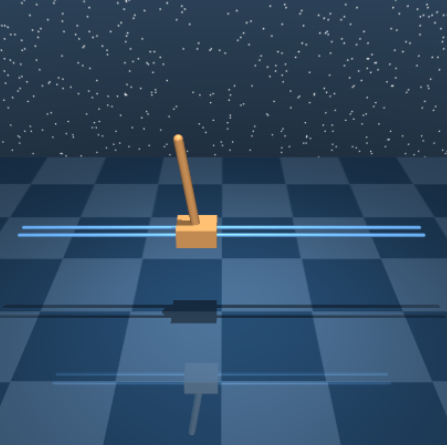}
  \end{center}
  \vspace{-4mm}
  \caption{The cartpole environment.}
  \vspace{-5mm}
  \label{figCartpole}
\end{wrapfigure}

The Cartpole swing-up environment is characterized by a five-dimensional state space and a single-dimensional action space. The action in this environment consists of applying a horizontal force to the base of the cart, with the force ranging from -1 to 1. The task horizon is set to 1000. The reward function is designed to encourage the system to remain upright and centered, with small action and velocity values. The observables for this environment include the x-coordinate, the cosine and sine of the pole angle, the x velocity, and the pole's angular velocity. These parameters collectively define the dynamics and objectives of the Cartpole swing-up task.

\subsection{The Offline cartpole datasets}
\label{appendixdatasets}

In this section, we introduce the different datasets that are used to evaluate the multi-timestep models on the \textit{Cartpole (swingup)} environment. These datasets are collected using some \textit{behaviour policies} that are considered unknown to the agent (the combination of the world model and the actor that is using it for planning). All the datasets consist of $50$ episodes ($50k$ steps) split randomly as following: $36$ episodes for training, $4$ episodes for validation, and the remaining $10$ episodes are used for testing.

\begin{figure}[ht]
\begin{center}
   \includegraphics[width=.95\linewidth]{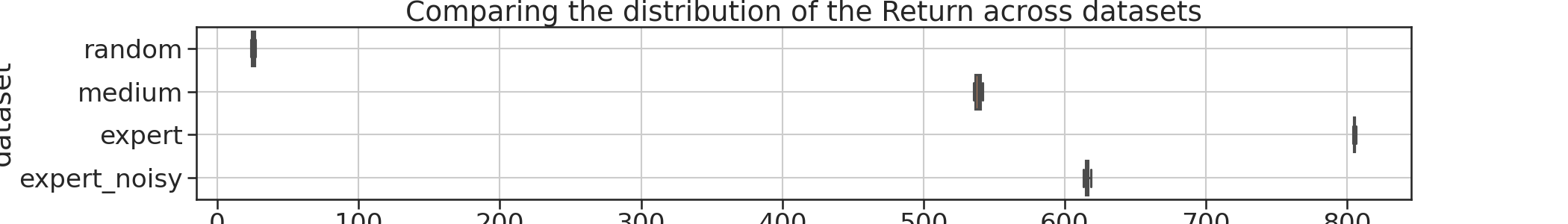}
\end{center}
\caption{A comparison of the distribution of returns across the considered datasets.}
\label{figDataReturn}
\end{figure}

We dispose of four datasets:

\begin{itemize}
    \item \textbf{random.} This dataset is collected using a random policy (a policy that samples actions uniformly from the action space).

    \item \textbf{medium.} This is an intermediate between the two. It is collected by an unstable SAC agent that sometimes reaches near-optimal behaviour and sometimes doesn't manage to.
    
    \item \textbf{expert.} This is the full learning trace of an \textit{expert}-policy (a Soft Actor-Critic -SAC- that is trained on an autoregressive mixture density network until convergence).

    \item \textbf{expert\_noisy.} This is the full learning trace of the \textbf{expert} policy in the noisy Cartpole environment.
\end{itemize}

\begin{figure}[!h]
\begin{center}
    \includegraphics[width=1.0\columnwidth]{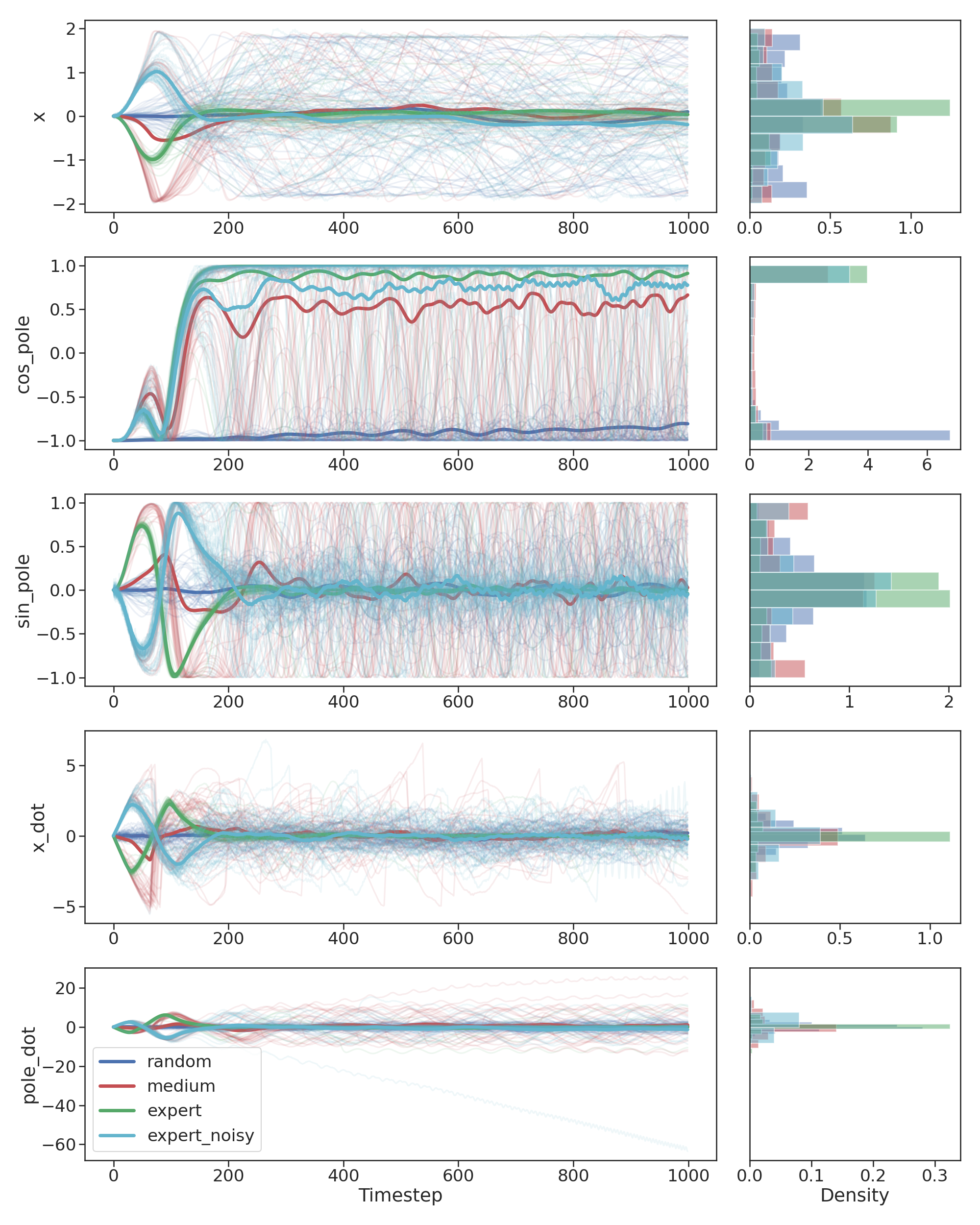}
\end{center}
\caption{A comparison of the distribution of the state observables across the considered datasets.}
\label{figDataVar}
\end{figure}

To understand the differences between these datasets we propose to analyze the variance of each state dimension of the Cartpole observables. We can see from Figure \ref{figDataVar} that the \textbf{expert} dataset has very little variance as it converges quickly (around episode $8$) and all the remaining episodes are almost identical. Similarly, the \textbf{random} dataset has little variance as it keeps exploring the same region around the initial state. The \textbf{medium} dataset, and naturally the \textbf{expert\_noisy} dataset are the most diverse among the four, and we believe it would be insightful to evaluate the multi-timestep models against these different challenges, both in terms of the predictive error, and the final return of the underlying agents.

\section{Implementation details}
\label{appendixsecImpDetails}

For all the models, we use a neural network composed of a common number of hidden layers and two output heads (with \textit{Tanh} activation functions) for the mean and standard deviation of the learned probabilistic dynamics (The standard deviation is fixed when we want to use the MSE loss). We use batch normalization \citep{Ioffe2015}, Dropout layers \citep{Srivastava2014} ($p=10\%$), and set the learning rate of the Adam optimizer \citep{Diederik2015} to $0.001$, the batch size to $64$, the number of common layers to $2$, and the number of hidden units to $256$ based on a hyperparameter search executed using the RAMP framework \citep{Kegl2018}. The evaluation metric of the hyperparameter optimization is the aggregated validation R2 score across all the offline datasets.

For the offline and iterated-batch RL experiments, we use SAC agents from the open-source library StableBaselines3 \citep{Raffin2021} while keeping its default hyperparameters. In the offline setting, we train the SAC agents for $500k$ steps on a fixed model, and evaluate them by rolling-out an episode in the real environment. In the iterated-batch setting, the episodes generated from the evaluation of SAC are added to the current data buffer and used to retrain the model from scratch. For both these setups, we modify the dynamics models to predict the state difference rather than the raw next state: $s_{t+1} \rLeadsto \hat{p}(s_t, a_t) + s_t$. This simple technique has been proved to improve the results of MBRL algorithms \citep{Chua2018}.

\section{Gradient computation}
\label{AppSecGrad}

In this section we want to compute the analytic derivative of the generalized loss: $L_\balpha\big(\bs_{t+1:t+h+1},\hat{p}^{1:h}(s_t, \ba_{\cdot})\big) = \sum_{j=1}^h \alpha_j L\big(s_{t+j},\hat{p}^j(s_t, \ba_{t:t+j})\big)$ with respect to a model parameter $\theta$. For simplicity we denote the j-step loss $L^{(j)} = L\big(s_{t+j},\hat{p}^j(s_t, \ba_{t:t+j})\big)$ and omit actions from the j-step prediction function: $ \hat{p}^j(s_t) $. Under the assumption of a single-dimensional state and MSE loss, we derive the formula of $\frac{d}{d \theta} L^{(j)}$ for $j \in \{1,\hdots,h\}$:

\begin{equation} \label{eq2}
\begin{split}
\frac{d}{d \theta} [ L^{(j)} ]& = \frac{d}{d \theta} [ (\hat{p}_\theta^j (s_t) - s_{t+j})^2 ]\\
 & = 2 \frac{d}{d \theta} [\hat{p}_\theta^j (s_t)] (\hat{p}_\theta^j (s_t) - s_{t+j})\\
 & = 2 \frac{d}{d \theta} [\hat{p}_\theta(\hat{p}_\theta^{j-1}(s_t))] (\hat{p}_\theta^j (s_t) - s_{t+j})\\
 & = 2 \frac{d}{d \theta} [\hat{p}_\theta^{j-1} (s_t)] \frac{d}{d \theta} [\hat{p}_\theta]\big(\hat{p}_\theta^{j-1}(s_t)\big) (\hat{p}_\theta^j (s_t) - s_{t+j})\\
 & = 2 \left( \prod_{i=0}^{j-1} \frac{d}{d \theta} [\hat{p}_\theta]\big(\hat{p}_\theta^{i}(s_t)\big) \right) (\hat{p}_\theta^j (s_t) - s_{t+j}) \hspace{.4cm} \text{(by recursion)}
\end{split}
\end{equation}

From this we can compute the gradient of the loss $\frac{d}{d \theta} L_\balpha$:

\begin{equation} \label{eq3}
\begin{split}
\frac{d}{d \theta} \mathcal{L} & = \sum_{j=1}^n \alpha_j \frac{d}{d \theta} L^{(j)}\\
& = 2 \sum_{j=1}^n \alpha_j \left( \prod_{i=0}^{j-1} \frac{d}{d \theta} [\hat{p}_\theta]\big(\hat{p}_\theta^{i}(s_t)\big) \right) (\hat{p}_\theta^j (s_t) - s_{t+j})
\end{split}
\end{equation}



We start by observing that the terms featuring in the product are cumulative as we go further in the horizon. We thus compute the ratio between two consecutive loss terms $L^{(j-1)}$ and $\mathcal{L}^{(j)}$:

\begin{equation} \label{eq4}
\begin{split}
\frac{\frac{d}{d \theta} L^{(j)}}{\frac{d}{d \theta} L^{(j-1)}} & = \frac{2 \left( \prod_{i=0}^{j-1} \frac{d}{d \theta} [\hat{p}_\theta]\big(\hat{p}_\theta^{i}(s_t)\big) \right) (\hat{p}_\theta^j (s_t) - s_{t+j})}{2 \left( \prod_{i=0}^{j-2} \frac{d}{d \theta} [\hat{p}_\theta]\big(\hat{p}_\theta^{i}(s_t)\big) \right) (\hat{p}_\theta^{j-1} (s_t) - s_{t+j-1})}\\
& = \frac{d}{d \theta} [\hat{p}_\theta]\big(\hat{p}_\theta^{j-1}(s_t)\big) \frac{(\hat{p}_\theta^j (s_t) - s_{t+j})}{(\hat{p}_\theta^{j-1} (s_t) - s_{t+j-1})}
\end{split}
\end{equation}

Given equation \ref{eq4}, we can write a formula for $\frac{d}{d \theta} L_\balpha$ that only depends on the prediction one-step ahead and its gradient.  We denote the error ratio between horizons $k$ and $l$: $Err_{(k,l)} = \frac{(\hat{p}_\theta^l (s_t) - s_{t+l})}{(\hat{p}_\theta^{k} (s_t) - s_{t+k})}$; And the gradient of the prediction function evaluated at the $k$-th horizon: $G_{k} = \frac{d}{d \theta} [\hat{p}_\theta]\big(\hat{p}_\theta^{k}(s_t)\big)$. We use the convention that $G_0=1$ and $Err_{(0,1)}=1$.

Let's refactor the expression of the gradient $\frac{d}{d \theta} L_\balpha$ in equation \ref{eq3} using the ratio from equation \ref{eq4}:

\begin{equation} \label{eq5}
\begin{split}
\frac{d}{d \theta} L_\balpha & = \sum_{j=1}^n \alpha_j \frac{d}{d \theta} L^{(j)}\\
& = \alpha_1 \frac{d}{d \theta} L^{(1)} + \sum_{j=2}^h \alpha_j \frac{d}{d \theta} [\hat{p}_\theta]\big(\hat{p}_\theta^{j-1}(s_t)\big) \frac{(\hat{p}_\theta^j (s_t) - s_{t+j})}{(\hat{p}_\theta^{j-1} (s_t) - s_{t+j-1})} L^{(j-1)}\\
& = \alpha_1 \frac{d}{d \theta} L^{(1)} + \sum_{j=2}^h \alpha_j G_{j-1} Err_{(j-1,j)} L^{(j-1)}\\
& = \alpha_1 \frac{d}{d \theta} L^{(1)} + \sum_{j=1}^{h-1} \alpha_{j+1} G_{j} Err_{(j,j+1)} L^{(j)}\\
& = \alpha_1 \frac{d}{d \theta} L^{(1)} + \alpha^2 G_{1} Err_{(1,2)} \frac{d}{d \theta} L^{(1)} + \sum_{j=2}^n \alpha_j G_{j-1} Err_{(j-1,j)} L^{(j-1)}\\
& = \sum_{j=1}^{h} \alpha_j \left( \prod_{i=0}^{j-1} G_i \right) \left( \prod_{i=0}^{j-1} Err_{(i,i+1)} \right) \frac{d}{d \theta} L^{(1)} \hspace{.5cm} \text{(by recursion)}\\
& = \sum_{j=1}^{h} \alpha_j \left( \prod_{i=0}^{j-1} G_i \right) \left( Err_{(1,h)} \right) \frac{d}{d \theta} L^{(1)} \hspace{.5cm} \text{(Elements simplify by definition of $Err_{(i,i+1)}$)}
\end{split}
\end{equation} 





This formulation expresses the generalized loss derivative as a linear function of the derivative of the loss one-step ahead. 
The latter is relatively cheap and can be computed with one forward (and backward) pass through the model. 
Consequently, we can think of approximation schemes to reduce the computational burden $O(n \times Cost(L^{(1)}))$ that comes with computing the full derivative: $\frac{d}{d \theta} L_\balpha$. 
We leave the exploration of this idea to a future follow-up work.

\end{document}